\documentclass[twocolumn]{fairmeta}
\usepackage{amsmath,amssymb}
\usepackage{adjustbox}
\usepackage{enumitem}
\usepackage{algorithm}
\usepackage{algpseudocode}
\usepackage{xspace}
\usepackage{tabularx}
\usepackage{needspace}
\newcommand{\FfourR}{F4R\xspace}
\newcommand{\NA}{--}
\newcolumntype{Y}{>{\raggedright\arraybackslash}X}

\title{F4R: Failure-Driven Recognition, Reconstruction, Refinement, and Redeployment for Continual Robot Self-Improvement}
\renewcommand{\authorformat}[2][]{\mbox{{\sffamily\bfseries #2$^{#1}$}}}
\author[1,*,\S]{Zhuoyuan Yu}
\author[3,*]{Jiacheng Wang}
\author[2,*]{Tianle Liu}
\author[*]{Yihua Ren}
\author[3]{Peng Yu}
\author[2]{Chen Bai}
\author[2,\ddagger]{Ziheng Zhang}
\author[2]{Yufei Jia}
\author[1]{Jindou Jia}
\author[1]{Yuhang Zhang}
\author[2]{Xinrui Zhang}
\author[1]{Shang Yujing}
\author[2]{Yuxiang Chen}
\author[1,\dagger]{Chuhao Zhou}
\author[2,\dagger]{Tiancai Wang}
\author[1,\dagger]{Jianfei Yang}
\affiliation[1]{Nanyang Technological University}
\affiliation[2]{Dexmal}
\affiliation[3]{Xi'an Jiaotong University}
\renewcommand{\contributionlist}{{\small
  $^{*}$Equal contribution\quad
  $^{\dagger}$Corresponding authors\quad
  $^{\ddagger}$Project leader\par
  $^{\S}$Work done during interning at Dexmal.
}}
\abstract{The real-world performance of current vision-language-action models is fundamentally constrained by the limited coverage of expert demonstrations and their insufficient understanding of physical interactions. A common remedy is to collect additional real-world demonstrations of newly encountered failures. However, this process is costly, inefficient, potentially unsafe, and difficult to scale. To address this challenge, we propose Failure for Rising (\textbf{F4R}), a \textbf{failure-driven real-to-sim-to-real closed-loop} learning framework that converts real-world failures into targeted policy improvement. F4R first uses an agent to automatically identify and diagnose failures from rollouts. It reconstructs each failure as an interactive, object-centric table-top environment that preserves the task-relevant spatial and physical conditions. The policy is then refined through failure-conditioned sim-real co-training followed by targeted reinforcement learning in the reconstructed environments. The improved policy is subsequently redeployed, while newly observed failures are continuously fed back into the next reconstruction and learning cycle. Real-world evaluations on four manipulation tasks show that F4R achieves 93.75\% In-Distribution and 90.0\% Out-of-Distribution (OOD) success, outperforming the budget-matched Targeted BC baseline by 18.75 percentage points under OOD conditions without collecting additional real-world corrective demonstrations.

Project Website: \url{https://yuj0e.github.io/F4R_Website} }
\hypersetup{
pdftitle={F4R: Failure-Driven Recognition, Reconstruction, Refinement, and Redeployment for Continual Robot Self-Improvement},
pdfauthor={Zhuoyuan Yu, Jiacheng Wang, Tianle Liu, Yihua Ren, Peng Yu, Chen Bai, Ziheng Zhang, Yufei Jia, Jindou Jia, Yuhang Zhang, Chuhao Zhou, Xinrui Zhang, Shang Yujing, Yuxiang Chen, Tiancai Wang, Jianfei Yang},
pdfsubject={Main paper and supplementary material}
}
\begin{document}
\raggedbottom
\maketitle
\section{Introduction}

Recent advances in robot learning have substantially expanded robots' ability to perform diverse manipulation tasks. Modern vision-language-action models are trained on large and diverse demonstration datasets~\citep{kim2025openvla,black2024pi0}. This scale enables them to learn broad visuomotor priors and generalize across tasks, objects, scenes, and robot embodiments. Nevertheless, strong performance on established benchmarks does not necessarily translate into reliable real-world deployment. Current policies remain brittle under Out-Of-Distribution (OOD) settings, e.g., unfamiliar object configurations, changes in camera viewpoint or robot initialization, and the presence of distracting objects or visual clutter~\citep{fei2026liberoplus,lee2025molmoact}.
Such variations can silently disrupt perception, spatial reasoning, and closed-loop action execution, ultimately leading to task failures. 
More importantly, these failures tend to arise in long-tailed scenarios that are underrepresented or even entirely absent in the original training data, exposing a persistent gap between training data and real-world conditions.

\begin{figure}[!t]
  \centering
  \includegraphics[width=\columnwidth]{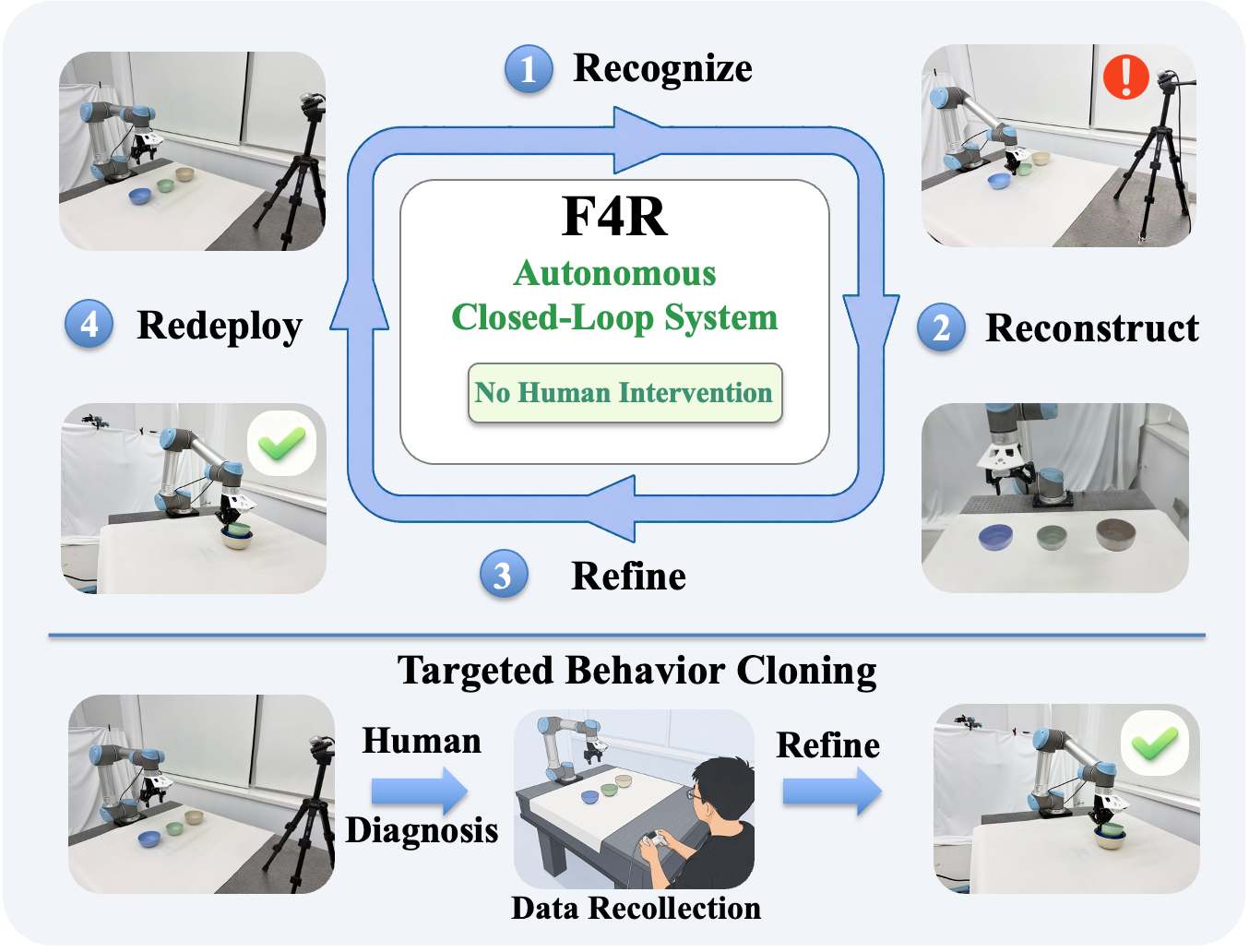}
  \caption{\textbf{Comparison of policy-improvement paradigms.} F4R forms an autonomous closed loop that recognizes deployment failures, reconstructs them in simulation, refines the policy, and redeploys it without human intervention. In contrast, targeted behavior cloning follows an open-loop pipeline that requires manual diagnosis and repeated real-world data recollection, resulting in substantial human effort and limited scalability.}
  \label{fig:cover}
\end{figure}

A straightforward way to improve coverage of long-tail scenarios is to collect more diverse real-world demonstrations. 
However, real-world data collection requires substantial human effort, hardware resources, and operational coordination, making it impractical to exhaustively capture rare failure cases~\citep{khazatsky2024droid}. 
The real-to-sim-to-real paradigm offers a more scalable alternative by using real-world data to construct simulation environments, where tasks, scenes, and initial conditions can be systematically diversified for policy improvement~\citep{mandlekar2023mimicgen,nasiriany2024robocasa}.

However, the effectiveness of real-to-sim-to-real learning depends on how the simulation environments are constructed. Manually designed scenarios or broad randomization may increase diversity, but they are not necessarily aligned with the failure distribution encountered in real-world deployment~\citep{lin2026failsafe}. 
As a result, the added diversity may improve overall performance, yet it does not necessarily provide targeted supervision signals for the recurring failure modes observed during real-world deployment.
Our key insight is that each real-world failure identifies a specific set of conditions under which the current policy lacks robustness. 
Reconstructing these conditions in interactive simulation turns each real-world failure into a targeted and reusable training scenario, allowing the robot to revisit hard cases and explore corrective behaviors without collecting additional costly and unsafe real-world demonstrations.

Building on this insight, we propose \textbf{F4R}, a failure-driven real-to-sim-to-real framework that iteratively refines robot policies using failures observed during deployment. 
F4R employs a Vision-Language Model (VLM)-based diagnostic agent to identify failed executions and infer their underlying causes from deployment videos, and then reconstructs the corresponding objects and spatial configurations in interactive simulation.
Since each reconstructed scene captures only one instance of a failure mode, F4R applies failure-aware randomization to generate a diverse set of failure-relevant configurations around the reconstructed scene.
Using successful real-world demonstrations as seeds, F4R uses MimicGen to adapt their object-centric motion segments to the randomized reconstructed scenes and retains successful simulated rollouts for sim-real co-training.
Following sim-real co-training, F4R further refines the policy through targeted reinforcement learning in the same environments before redeployment, allowing newly observed failures to drive subsequent refinement cycles. 
With this iterative refinement process, F4R achieves a 90.0\% OOD success rate, outperforming the budget-matched Targeted BC baseline by 18.75 percentage points without additional real-world corrective demonstrations.

Our contributions are summarized as follows:
\begin{enumerate}
    \item We develop a failure-guided reconstruction pipeline that converts real-world failures into geometrically consistent, photorealistic, and interactive simulation environments.

    \item We propose a two-stage refinement strategy that combines failure-relevant sim-real co-training with targeted reinforcement learning, allowing newly observed failures to drive subsequent refinement cycles.

    \item The components are integrated into F4R, a real-to-sim-to-real framework that iteratively turns real-world failures into reusable training assets. Extensive experiments demonstrate its effectiveness in policy improvement without additional real-world corrective demonstrations.

\end{enumerate}

\section{Related Work}

\subsection{Real-to-Sim Reconstruction for Robot Learning}

Grounding simulation environments in real-world observations provides a promising approach to narrowing the sim-to-real gap. RialTo~\citep{villasevil2024reconciling} reconstructs digital twins from real-world scans for RL-based policy adaptation. Recent studies increasingly leverage 3D Gaussian Splatting to improve visual fidelity. RL-GSBridge~\citep{wu2025rl} couples Gaussian representations with simulation meshes, while RoboGSim~\citep{li2024robogsim} and GS-Playground~\citep{jia2026gsplayground} integrate scene reconstruction, composition, physics, and data generation. RoboSplat~\citep{YangS-RSS-25} enables scene editing for data diversification, whereas EmbodieDreamer~\citep{wang2025embodiedreamer} and RoboSimGS~\citep{11425018} further improve visual and physical alignment. In parallel, URDFormer~\citep{Chen-RSS-24} and Scalable Real2Sim~\citep{11246653} automate the recovery of simulation-ready geometry and physical properties.

Most existing pipelines, however, remain scene- or task-driven: the target environment is typically selected independently of the policy's observed deployment failures. Consequently, the generated simulation data may increase overall diversity but are not explicitly concentrated on the hardest conditions under which the deployed policy is least robust. In contrast, F4R uses deployment failures to determine both what should be reconstructed and how each reconstructed scene should be expanded, thereby forming a failure-centered training distribution for targeted policy refinement.

\subsection{VLA Adaptation via Sim-Real Co-Training and RL}

Supervised fine-tuning is widely used to adapt generalist robot policies, but its performance is constrained by the coverage of available demonstrations. Sim-Real Co-training~\citep{maddukuri2025sim} broadens this coverage by jointly training on real and simulated data, while task-relevant representation alignment improves cross-domain transfer~\citep{cheng2025domainadaptation}. Nevertheless, these approaches still rely on simulated demonstrations that sufficiently cover deployment-relevant conditions.

Reinforcement learning complements supervised training by improving policies through interaction. ReinboT~\citep{zhang2025reinbot} introduces return maximization into offline VLA training, while ConRFT~\citep{chen2025conrft} combines offline value learning with online refinement. SimpleVLA-RL~\citep{li2026simplevlarl} scales outcome-based online RL through VLA-specific sampling and parallel interaction, and ReinFlow~\citep{zhang2026reinflow} and FPO~\citep{lyu2025reinforcement} extend RL to flow-based policies. RLinf-Co~\citep{shi2026beyond} combines sim-real co-training with simulation RL and real-data supervision. However, these methods generally optimize over predefined distributions rather than using deployment failures to determine where refinement is most needed.

Recent systems exploit execution feedback more explicitly. ASPIRE~\citep{lu2026aspire} diagnoses failures and repairs code-as-policy programs, consolidating validated solutions into a reusable skill library. TwinRL~\citep{xu2026twinrl} reconstructs a workspace-level digital twin to expand SFT coverage, initialize simulation RL, and guide human-in-the-loop real-world refinement. In contrast, F4R uses deployment failures to jointly determine what should be reconstructed and where training should be concentrated, forming a closed loop that converts newly observed failures into reusable simulation environments and targeted policy updates.

\section{Method}

\begin{figure*}[t] 
  \centering
  \includegraphics[width=\textwidth]{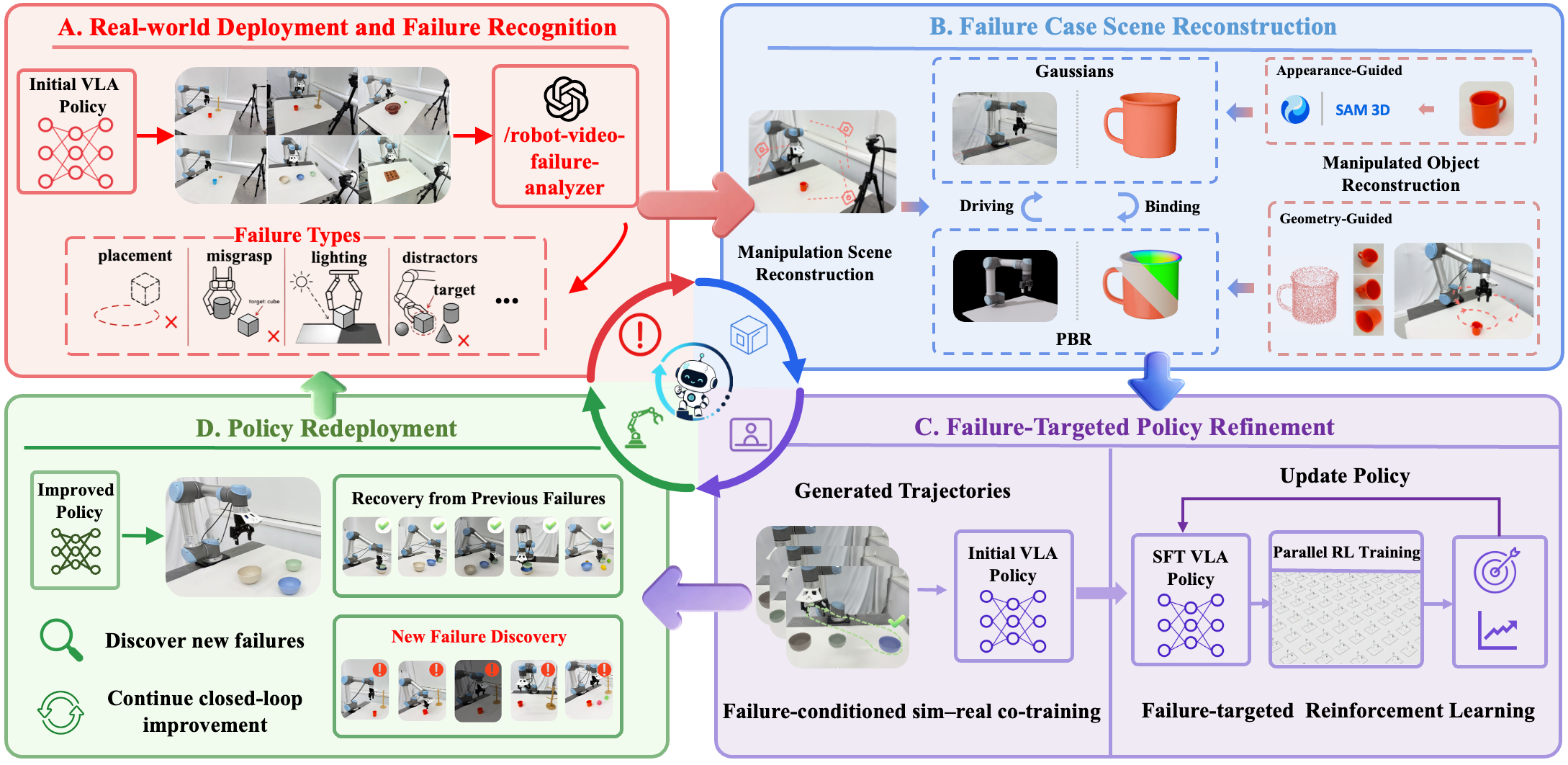}
  \caption{\textbf{Overview of the F4R architecture.} (A) An initial policy is deployed in the real world to collect failure trajectories and task-relevant observations. (B) F4R reconstructs the failure scene in two stages. The resulting PBR meshes drive physical simulation, while the aligned Gaussians enable photorealistic rendering. (C) The policy first acquires corrective behaviors through failure-conditioned sim-real co-training and then improves them through targeted RL in the reconstructed environments. (D) The refined policy is redeployed and discovers new failure cases, closing the real-to-sim-to-real learning loop.}
  \label{fig:overview}
\end{figure*}

F4R is an iterative real-to-sim-to-real framework that turns
real-world failures into reusable training assets. As illustrated
in Figure.~\ref{fig:overview}, at refinement
round $r$, the current policy $\pi_{\theta}^{(r)}$ is deployed in
the real world to collect failed rollouts. F4R then (i) identifies
the earliest failed subtask and the scene factors associated with
it, (ii) reconstructs these task-relevant conditions in simulation
and expands them through failure-aware randomization, (iii)
generates successful corrective demonstrations and refines the
policy through sim-real co-training followed by targeted
reinforcement learning, and (iv) redeploys the updated policy to
initiate the next round.

\subsection{Problem Formulation}

Following~\citep{chi2023diffusion}, we denote $\mathcal{D}_{\mathrm{real}}$ as the original real-world demonstration set and the policy at refinement round $r$ by $\pi_{\theta}^{(r)}$.
At time step $t$, the policy receives a visual observation $o_t$,
robot state $s_t$, and language instruction $\ell$. Let
$c_t=(o_t,s_t,\ell)$ denote the policy context at time $t$. Given $c_t$, the policy predicts an action chunk
\begin{equation}
    \mathbf{a}_{t:t+H-1}
    \sim
    \pi_{\theta}^{(r)}\!\left(\cdot\mid c_t\right),
    \label{eq:f4r_policy}
\end{equation}
where $H$ is the action horizon. A deployment rollout is denoted by
$\tau_i=\{(c_{i,t},a_{i,t})\}_{t=1}^{T_i}$, with binary task
outcome $\mathcal{S}(\tau_i)\in\{0,1\}$. The failures observed at
round $r$ form
$\mathcal{D}_{\mathrm{fail}}^{(r)}
=\{\tau_i\mid\mathcal{S}(\tau_i)=0\}$, and
$N_r=|\mathcal{D}_{\mathrm{fail}}^{(r)}|$. The goal is to improve
the policy around these observed failure scenarios while
preserving the capabilities supported by
$\mathcal{D}_{\mathrm{real}}$.

\subsection{Failure Recognition and Diagnosis}

For each failed rollout $\tau_i$ of $T_i$ frames, F4R receives
synchronized third-person and wrist-view image streams
$\mathbf{I}_i^{g}=\{I_{i,t}^{g}\}_{t=1}^{T_i}$ and
$\mathbf{I}_i^{w}=\{I_{i,t}^{w}\}_{t=1}^{T_i}$, together with
the task instruction $\ell_i$. A lightweight temporal proposal
module $\mathcal{P}$ identifies a small set of interaction-centered
intervals from the two video streams. Each interval is summarized
as a temporally ordered multi-view evidence board:
\begin{equation}
\begin{aligned}
    \mathcal{W}_i
    &=
    \mathcal{P}\!\left(
        \mathbf{I}_i^{g},
        \mathbf{I}_i^{w}
    \right)
    =
    \left\{
        [s_{i,k},e_{i,k}]
    \right\}_{k=1}^{K_i},
    \\
    \mathcal{B}_{i,k}
    &=
    \operatorname{Board}\!\left(
        \left\{
            \left(I_{i,t}^{g},I_{i,t}^{w},t\right)
        \right\}_{t=s_{i,k}}^{e_{i,k}}
    \right).
\end{aligned}
\label{eq:f4r_temporal_proposal}
\end{equation}
Here, $[s_{i,k},e_{i,k}]$ denotes the $k$-th proposed
temporal interval, and $\operatorname{Board}(\cdot)$ arranges
synchronized third-person and wrist-view frames, together with
their timestamps, into a temporally ordered visual board. The
intervals are ordered chronologically, while an additional state
board $\mathcal{B}_{i}^{\mathrm{state}}$ summarizes the initial
and terminal scene states.

A VLM-based diagnostic agent $\mathcal{A}_{\phi}$ examines the
resulting evidence boards in temporal order and identifies the
earliest visually supported failure, producing a structured
failure record
\begin{equation}
\begin{aligned}
    \mathcal{F}_i
    &=\mathcal{A}_{\phi}\!\left(
        \{\mathcal{B}_{i,k}\}_{k=1}^{K_i},
        \mathcal{B}_{i}^{\mathrm{state}},\ell_i
    \right)\\
    &=\left(t_i^{\star},\mathcal{O}_i,\sigma_i,
        \mathcal{Z}_i,\mathcal{E}_i,\kappa_i\right),
\end{aligned}
\label{eq:f4r_failure_record}
\end{equation}
where $t_i^{\star}$ is the earliest failure time,
$\mathcal{O}_i$ contains the task-relevant objects,
$\sigma_i$ describes the failed subtask and failure type,
$\mathcal{Z}_i$ specifies the failure-relevant scene factors,
$\mathcal{E}_i$ contains frame-grounded evidence, and
$\kappa_i$ denotes the diagnosis confidence. The resulting
record determines which scene components should be reconstructed
and which variables should be randomized. Additional
implementation details on temporal proposal generation, board
construction, and diagnostic prompting are provided in
Appendix.

\subsection{Failure Case Reconstruction}

Conditioned on the failure record $\mathcal{F}_i$, F4R
constructs a nominal interactive environment
$\widehat{\mathcal{M}}_i$ that preserves the task-relevant
objects $\mathcal{O}_i$, their spatial configuration, and the
failure-associated conditions $\mathcal{Z}_i$. Reconstruction
is performed at two levels: a shared manipulation scene
recovers the static workspace, while reusable object assets
recover the geometry, appearance, and optional articulation
of the manipulated objects.

\paragraph{Manipulation scene reconstruction.}
As a one-time setup, a short RGB sequence of the workspace
is captured with a smartphone and reconstructed using 3D
Gaussian Splatting. The reconstructed workspace is registered
to the simulator world frame and integrated with the robot
and camera models in Isaac Sim. We denote the resulting
shared scene asset by $\mathsf{S}$. It provides a
photorealistic representation of the static workspace and is
reused across failure cases and refinement rounds. This
capture is the only human-assisted stage of F4R, while
failure diagnosis, object reconstruction, scene adaptation,
and policy refinement proceed automatically.

\paragraph{Manipulated object reconstruction.}
For each task-relevant object $o\in\mathcal{O}_i$, the robot
autonomously acquires multi-view RGB-D observations
$\mathcal{V}_o$ using its wrist camera. The observations are
stored in a COLMAP-style format~\citep{schoenberger2016sfm},
including calibrated RGB-D frames, camera intrinsics and poses,
and optional object masks. The reconstruction pipeline is
summarized as
\begin{equation}
\begin{aligned}
    \mathcal{V}_o
    &\rightarrow
    g_o^{\mathrm{pcd}}
    \rightarrow
    g_o^{\mathrm{mesh}}
    \rightarrow
    \bar{g}_o^{\mathrm{mesh}}\rightarrow
    g_o^{\mathrm{PBR}}
    \rightarrow
    g_o^{\mathrm{GS}} .
\end{aligned}
\label{eq:f4r_object_reconstruction}
\end{equation}

Specifically, depth refinement and TSDF
fusion~\citep{lingbot-depth2026,curless1996volumetric} produce
the metric point cloud $g_o^{\mathrm{pcd}}$.
ShapeR~\citep{siddiqui2026shaper} completes the geometry into
$g_o^{\mathrm{mesh}}$, which is regularized into a watertight
mesh $\bar{g}_o^{\mathrm{mesh}}$. For articulated objects, a
VLM together with P3-SAM~\citep{ma2025p3} recovers the
part-level kinematic structure $\mathcal{K}_o$.
Hunyuan3D~\citep{hunyuan3d2025hunyuan3d} synthesizes PBR
materials to obtain $g_o^{\mathrm{PBR}}$, whose renderings are
used to optimize the aligned Gaussian representation
$g_o^{\mathrm{GS}}$. The object asset is represented as
$\mathsf{A}_o=(g_o^{\mathrm{PBR}},g_o^{\mathrm{GS}},
\mathcal{K}_o)$, supporting physical interaction,
photorealistic rendering, and optional articulation,
respectively. 
Regardless of the reconstruction route,
the diagnosed factors $\mathcal{Z}_i$ are grounded using the
visual evidence $\mathcal{E}_i$ into a nominal simulator
parameter vector $\widehat{\boldsymbol{\xi}}_i$, which specifies
the object poses and articulation states, robot initialization,
camera configuration, appearance, and physical properties.
The shared scene asset $\mathsf{S}$ and object assets
$\{\mathsf{A}_o\}_{o\in\mathcal{O}_i}$ are then assembled as
\begin{equation}
    \widehat{\mathcal{M}}_i
    =
    \operatorname{Assemble}\!\left(
        \mathsf{S},
        \{\mathsf{A}_o\}_{o\in\mathcal{O}_i};
        \widehat{\boldsymbol{\xi}}_i
    \right).
    \label{eq:f4r_environment_assembly}
\end{equation}
The resulting environment instantiates the task-relevant
conditions associated with failure $i$ and serves as the center
of subsequent failure-aware randomization.

\begin{table*}[!t]
    \centering
    \small
    \setlength{\tabcolsep}{4.2pt}
    \renewcommand{\arraystretch}{1.08}
    \resizebox{\textwidth}{!}{
    \begin{tabular}{ccccccccccc}
        \toprule

        \multirow[c]{2}{*}[-2ex]{Method}
        & \multicolumn{4}{c}{In Distribution}
        & \multirow[c]{2}{*}[-2ex]{ID Avg.}
        & \multicolumn{4}{c}{Out of Distribution}
        & \multirow[c]{2}{*}[-2ex]{OOD Avg.} \\

        \cmidrule(lr){2-5}
        \cmidrule(lr){7-10}

        & \shortstack[c]{Pick\\Fruits}
        & \shortstack[c]{Place Cup\\on Coaster}
        & \shortstack[c]{Stack\\Bowls}
        & \shortstack[c]{Place Block\\in Drawer}
        &
        & \shortstack[c]{Pick\\Fruits}
        & \shortstack[c]{Place Cup\\on Coaster}
        & \shortstack[c]{Stack\\Bowls}
        & \shortstack[c]{Place Block\\in Drawer}
        & \\

        \midrule

        Base
        & 70\% & 60\% & 50\% & 70\% & 62.5\%
        & 40\% & 35\% & 30\% & 0\% & 26.25\% \\

        Targeted BC
        & 100\% & 95\% & 80\% & 90\% & 91.25\%
        & 85\% & 80\% & 60\% & 60\% & 71.25\% \\

        RLinf-Co
        & 95\% & 85\% & 80\% & 90\% & 87.5\%
        & 90\% & 70\% & 70\% & 80\% & 77.5\% \\

        \textbf{F4R (Ours)}
& \textbf{100\%}
& \textbf{95\%}
& \textbf{85\%}
& \textbf{95\%}
& \textbf{93.75\%}
& \textbf{100\%}
& \textbf{95\%}
& \textbf{80\%}
& \textbf{85\%}
& \textbf{90\%} \\

        \bottomrule
    \end{tabular}
    }
    \caption{Success rates on four representative tasks under the original distribution and the deployment-derived failure distribution.}
    \label{tab:main_results}
\end{table*}

\subsection{Failure-Targeted Policy Refinement}

A reconstructed scene represents only one realization of a failure and may cause overfitting. F4R expands each scene into a local failure-centered distribution and refines the policy through sim-real co-training followed by targeted RL.

\paragraph{Failure-aware domain randomization.}
Let $\widehat{\mathcal{M}}_i$ denote the reconstructed simulation environment for failure $i$, and let $\xi_i^f$ denote its recovered parameter vector, including object poses, robot initialization, camera configuration, appearance, and physical properties. F4R samples a new parameter vector $\xi$ and constructs a randomized environment $\mathcal{M}_{i,\xi}$ as
\begin{equation}
    \xi\sim p_i^{F}\!\left(\xi\mid\xi_i^f,\mathcal{F}_i\right),
    \qquad
    \mathcal{M}_{i,\xi}
    =
    \mathrm{Rand}\!\left(\widehat{\mathcal{M}}_i;\xi\right),
    \label{eq:f4r_randomization}
\end{equation}
where the failure record $\mathcal{F}_i$ determines the randomized variables and their ranges. Failure-relevant variables are sampled around the recovered configuration, while unrelated variables remain fixed. This preserves task semantics while covering local variations around the observed failure.

The overall failure-centered distribution is
\begin{equation}
    p_F(\mathcal{M})
    =
    \frac{1}{|\mathcal{D}_{\mathrm{fail}}|}
    \sum_{i=1}^{|\mathcal{D}_{\mathrm{fail}}|}
    p_i^{F}\!\left(
    \mathcal{M}\mid\widehat{\mathcal{M}}_i,\mathcal{F}_i
    \right).
    \label{eq:f4r_distribution}
\end{equation}
It determines both where corrective demonstrations are generated and where RL interaction is concentrated.

\paragraph{Stage I: Failure-conditioned sim--real co-training.}
F4R uses successful trajectories from historical real-world rollouts as seed demonstrations for MimicGen~\citep{mandlekar2023mimicgen}. MimicGen adapts these demonstrations to environments sampled from $p_F(\mathcal{M})$, forming the failure-targeted dataset $\mathcal{D}_{\mathrm{sim}}^{F}$. If the available successful trajectories are insufficient to support MimicGen, AnyGrasp~\citep{fang2023anygrasp} is used to generate additional grasp proposals and supplement the seed demonstrations. The resulting simulated trajectories are co-trained with the original real-world data:
\begin{equation}
    \mathcal{L}_{\mathrm{CT}}(\theta)
    =
    \mathcal{L}_{\mathrm{SFT}}
    \left(\theta;\mathcal{D}_{\mathrm{real}}\right)
    +
    \lambda_{\mathrm{sim}}
    \mathcal{L}_{\mathrm{SFT}}
    \left(\theta;\mathcal{D}_{\mathrm{sim}}^{F}\right).
    \label{eq:f4r_cotraining}
\end{equation}
The simulated data provide failure-specific corrections, while the real data preserve existing capabilities and provide a stable initialization for RL.

\paragraph{Stage II: Failure-targeted reinforcement learning.}
Starting from the co-trained policy, $\theta_{\mathrm{RL}}^{(0)}=\theta_{\mathrm{CT}}$, F4R performs PPO in environments sampled from $p_F(\mathcal{M})$. It retains the original task reward without failure-specific reward shaping. An auxiliary SFT objective on the original real-world data is used during optimization:
\begin{equation}
    \mathcal{L}_{\mathrm{F4R}}(\theta)
    =
    \mathcal{L}_{\mathrm{PPO}}
    \left(\theta;p_F(\mathcal{M})\right)
    +
    \beta
    \mathcal{L}_{\mathrm{SFT}}
    \left(\theta;\mathcal{D}_{\mathrm{real}}\right).
    \label{eq:f4r_total_loss}
\end{equation}
RL improves closed-loop and contact-sensitive behaviors, while real-data supervision mitigates catastrophic forgetting.

\subsection{Closed-Loop Redeployment}

After refinement, the updated policy is redeployed and evaluated on both previous failures and new configurations. Failures that persist or emerge anew are incorporated into the next round of reconstruction and refinement. Through this iterative deployment--reconstruction--refinement process, F4R converts transient real-world failures into reusable training assets and progressively aligns policy improvement with the conditions encountered during actual operation.

\section{Experiments}
\label{sec:experiments}

We evaluate F4R through simulation and real-world experiments. Our evaluation addresses five questions. (1) Can F4R improve real-world performance on observed deployment failures? (2) Do the reconstructed environments preserve task-relevant behavioral difficulty and reduce real data requirements? (3) Does F4R generalize across different failure-inducing variations? (4) Does F4R support effective closed-loop policy improvement? (5) How do different parts affect the optimization and capability retention?

\subsection{Experimental Setup}
\label{sec:experimental_setup}

\begin{figure*}[!t]
    \centering
    \includegraphics[width=1\textwidth]{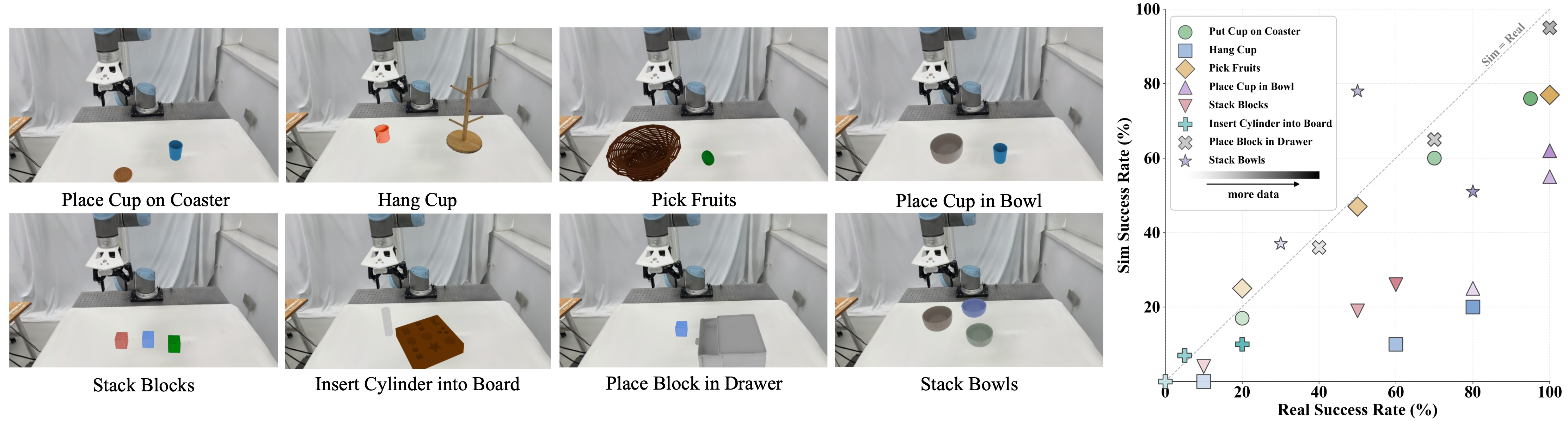}
    \caption{\textbf{Sim--real behavioral consistency.} Reconstructed environments for eight tabletop tasks and simulation versus real-world success rates for 24 policy checkpoints. Each marker represents one task and demonstration scale.}
    \label{fig:sim_real}
\end{figure*}

\paragraph{Tasks and Evaluation Protocol.}
We evaluate our method on eight tabletop manipulation tasks (Figure~\ref{fig:sim_real}) by comparing binary success rates. Four representative tasks are further selected for policy-refinement experiments. We define the In-Distribution (ID) setting as task configurations covered by the original data-collection distribution, and the Out-of-Distribution (OOD) setting as deployment failure conditions absent from that distribution. Real-world performance is measured over 20 trials in each setting, while simulation performance is evaluated using one rollout in each of 100 parallel environments. All methods use identical evaluation configurations and trial budgets.

\paragraph{Implementation Details.}
Our platform consists of a UR5 with a Robotiq 2F-85 gripper and two Intel RealSense D435i cameras: a fixed third-person camera and a wrist-mounted camera. We initialize the policy from $\pi_{0.5}$~\citep{intelligence2025pi05} and fine-tune LoRA adapters using AdamW. Supervised Fine-Tuning (SFT) is performed with FSDP on four NVIDIA H20 GPUs, using 32 samples per GPU. Policy refinement uses PPO with trainable LoRA adapters and a value head. Since H20 GPUs lack RT cores, two NVIDIA RTX 4090 GPUs run Isaac Lab and collect rollouts from 256 environments, while the H20 GPUs handle policy inference and optimization. The two sides communicate through a Ray cluster over SSH. PPO is trained for 200 update steps.

\subsection{Real-World Refinement on Deployment Failures}

Table~\ref{tab:main_results} evaluates F4R on failures encountered during real-world deployment. We compare against Targeted Behavior Cloning (Targeted BC), which collects corrective demonstrations from the real failure distribution, and RLinf-Co, which performs sim--real co-training followed by simulation RL. Targeted BC receives one hour of real-world data collection. RLinf-Co and F4R instead collect simulated experience for 10 minutes using 256 parallel environments, with the remaining 50 minutes accounting for reconstruction. In practice, scene reconstruction takes 25--30 minutes and object preparation takes 5--10 minutes. Importantly, scene reconstruction is a one-time setup. The reconstructed environment can be reused across subsequent failure-refinement cycles and only needs to be rebuilt when the physical workspace changes. Because task horizons vary, we match data-preparation time rather than trajectory count. RLinf-Co and F4R use identical architectures, training settings, and rollout budgets. Their only difference is the training distribution: RLinf-Co uses broad domain randomization, whereas F4R concentrates randomization around diagnosed failures.

Targeted BC achieves 91.25\% success on the original distribution and 71.25\% on the failure distribution, showing that corrective demonstrations are effective but require substantial real-world interaction. RLinf-Co improves failure-distribution success to 77.5\%. Its training loss also converges, suggesting that the remaining gap is not caused by insufficient optimization. Instead, broad randomization may enlarge the sim-to-real gap and dilute training around relevant failures. By concentrating both trajectory generation and RL interaction on failure-relevant variations, F4R achieves 93.75\% and 90.0\% success on the original and failure distributions, respectively. It outperforms Targeted BC and RLinf-Co by 18.75 and 12.5 percentage points on the failure distribution.

The benefit is particularly clear on the multi-stage \textit{Place Block in Drawer} task. Notably, when the drawer orientation differs from the original demonstrations, the base policy fails all closing attempts. This suggests that the supervised fine-tuned VLA model learns orientation-specific action patterns rather than transferable drawer kinematics. Targeted BC reaches 60\% success, while RLinf-Co and F4R achieve 80\% and 85\%, respectively. These results highlight the value of simulation interaction for acquiring contact-sensitive behaviors that are difficult to capture with fixed demonstrations.

\subsection{Reliability of Autonomous Failure Diagnosis}

We independently evaluate the reliability of F4R's diagnosis module on eight tasks. For each task, we collect ten rollout trajectories, yielding 80 trajectories in total. Each trajectory is manually annotated with its failure stage and underlying cause. GPT-5.5, equipped with our \texttt{robot-video-failure-analyzer} skill, then predicts both labels. A diagnosis is considered correct only when both predictions match the human annotations.

\begin{table}[htbp]
    \centering
    \small

    \setlength{\tabcolsep}{5pt}
    \renewcommand{\arraystretch}{1.08}

    \begin{adjustbox}{max width=\columnwidth}
    \begin{tabular}{lcc}
        \toprule
        Result & Trajectories & Rate \\
        \midrule
        Correct after one attempt
        & 69 / 80 & 86.25\% \\
        Correct within two attempts
        & 74 / 80 & 92.50\% \\
        \midrule
        Remaining: camera-viewpoint shift
        & 4 / 80 & 5.00\% \\
        Remaining: illumination change
        & 2 / 80 & 2.50\% \\
        \bottomrule
    \end{tabular}
    \end{adjustbox}
    \caption{\textbf{Autonomous failure-diagnosis accuracy.}
    Results on 80 manually annotated trajectories from eight tasks.}
    \label{tab:diagnosis_accuracy}
\end{table}

As shown in Table~\ref{tab:diagnosis_accuracy}, F4R correctly diagnoses 69 of 80 trajectories on the first attempt, achieving 86.25\% accuracy. A second attempt resolves five of the remaining eleven cases, increasing accuracy to 92.5\%. Among the six unresolved trajectories, four involve subtle camera-viewpoint shifts, and two involve illumination changes. These failures often require comparing the current observation with earlier frames to separate visual changes from actual task events. To assess compatibility with different VLM backends, we further evaluate several mainstream models on the public ViFailback dataset~\citep{zeng2025diagnose}. Without task-specific adaptation, the best-performing GPT model achieves nearly 70\% accuracy; complete results are provided in the appendix. These findings reveal a remaining limitation in reasoning over fine-grained temporal evidence, motivating the simulation-side validation and fallback mechanisms in F4R.

\subsection{Sim--Real Consistency and Data Efficiency}
\label{sec:sim_real_consistency}

A reconstructed environment is useful only if it preserves real-world behavioral trends. We evaluate this property across eight tabletop tasks with diverse objects and interaction complexities. For each task, we train policies using 10, 30, and 50 real-world demonstrations, yielding 24 checkpoints evaluated under matched configurations and success criteria in simulation and reality. This setup assesses whether reconstruction captures both task difficulty and performance gains from additional training data.

Figure~\ref{fig:sim_real} shows a strong positive correlation between simulation and real-world success rates (Pearson $r=0.7625$, $p=1.49\times10^{-5}$). Checkpoints trained with more demonstrations generally move toward the upper-right region, while the reconstructed environments preserve broad differences in task difficulty. Deviations from the diagonal reflect residual gaps in appearance, contact dynamics, and execution sensitivity, so simulation success is not an exact estimate of real-world performance. Nevertheless, the consistent ranking across tasks, checkpoints, and data scales supports using these environments as behavioral proxies.

\subsection{Robustness Across Failure Conditions}
\label{sec:failure_conditions}

We evaluate whether F4R generalizes beyond a single reconstructed condition using four heterogeneous deployment failure-inducing variations in \textit{Pick Fruits}. These variations cover geometric, visual, and clutter-related shifts.

\begin{table}[htbp]
    \centering
    \small
    \renewcommand{\arraystretch}{1.0}

    \begin{adjustbox}{max width=\columnwidth}
        \begin{tabular*}{1\columnwidth}{
            @{\extracolsep{\fill}}
            cccc
            @{}
        }
            \toprule
            Condition
            & Base
            & F4R w/o RL
            & \textbf{F4R} \\
            \midrule
            Object layout
            & 40\% & 100\% & \textbf{100\%} \\
            Illumination
            & 55\% & 80\% & \textbf{90\%} \\
            Background
            & 65\% & 100\% & \textbf{100\%} \\
            Distractors
            & 25\% & 55\% & \textbf{70\%} \\
            \bottomrule
        \end{tabular*}
    \end{adjustbox}
    \caption{\textbf{Robustness across failure conditions.}
    Real-world success rates on Pick Fruits under four variations.}
    \label{tab:failure_conditions}
\end{table}

As shown in Table~\ref{tab:failure_conditions}, F4R improves performance under every condition, increasing the average success rate from 46.25\% to 90.0\%. Co-training alone reaches 83.75\%, while targeted RL provides further gains under illumination and distractors. These consistent improvements demonstrate that F4R generalizes across diverse failure-inducing variations.

\subsection{Multi-Round Closed-Loop Refinement.}

\begin{figure}[htbp]
    \centering
    \includegraphics[width=\columnwidth]{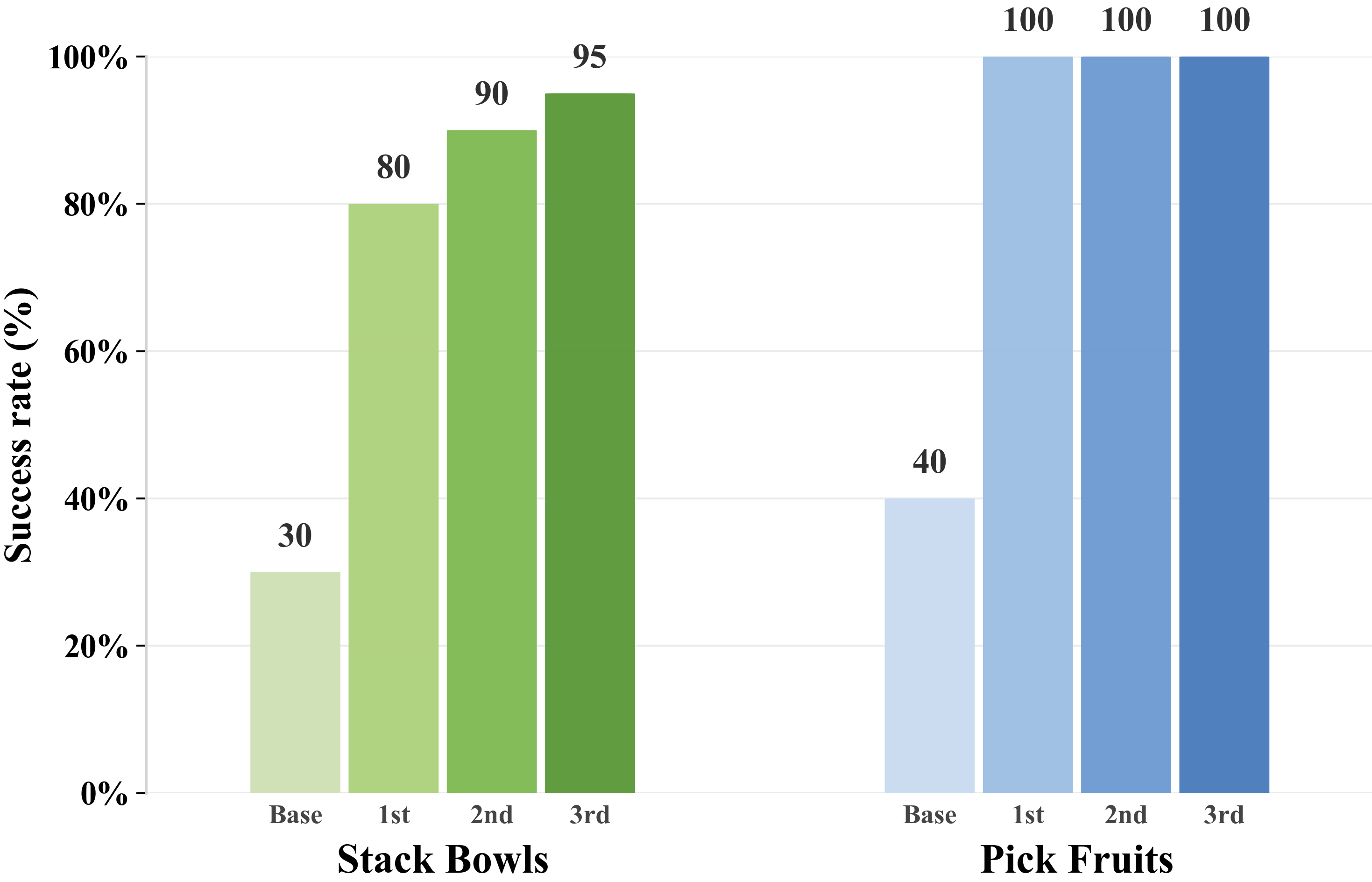}
    \caption{\textbf{Multi-round closed-loop policy refinement.}
Real-world success rates of the base policy and the refined policies after three consecutive F4R cycles.}
\label{fig:closed_loop}
\end{figure}

To evaluate multi-round closed-loop refinement, we conduct three consecutive refinement cycles on two tasks. After each cycle, the refined policy is redeployed, and newly observed failures are used to guide the next round of reconstruction and training. As shown in Figure~\ref{fig:closed_loop}, the success rate on \textit{Stack Bowls} progressively increases from 30\% to 80\%, 90\%, and 95\%. This gradual improvement suggests that successive cycles address residual failure modes that remain after earlier refinements. On \textit{Pick Fruits}, the first cycle raises performance from 40\% to 100\%, which is maintained throughout the next two cycles. The different convergence patterns indicate that F4R can support both gradual correction and rapid saturation, depending on task difficulty. Overall, these results show that F4R can continually address newly observed failures while preserving capabilities acquired in earlier rounds.

\subsection{Ablation Study}

\begin{table}[htbp]
    \centering
    \small
    \renewcommand{\arraystretch}{1.0}

    \begin{adjustbox}{max width=\columnwidth}
        \begin{tabular*}{1\columnwidth}{
            @{\extracolsep{\fill}}
            ccc
            @{}
        }
            \toprule
            Method & ID Avg. & OOD Avg. \\
            \midrule
            Base & 62.5\% & 26.25\% \\
            F4R w/o Failure Diag.  & 87.5\% & 77.5\% \\
            F4R w/o Real-Data & 85\% & 86.25\% \\
            \textbf{F4R} & \textbf{93.75\%} & \textbf{90\%} \\
            \bottomrule
        \end{tabular*}
    \end{adjustbox}
    \caption{\textbf{Ablation of failure diagnosis and real-data supervision.}
Average real-world success rates (\%) over four tasks under ID and OOD conditions.
``w/o Failure Diag.'' uses broad domain randomization, while
``w/o Real-Data'' removes real-data supervision from the SFT and RL stages.}
    \label{tab:ablation}
\end{table}

\paragraph{Effects of Failure Diagnosis and Real-Data Supervision.}
Table~\ref{tab:ablation} evaluates the two components. Replacing failure-aware randomization with broad randomization reduces ID and OOD success by 6.25 and 12.5 points, showing that diagnosis is particularly important for targeting OOD failures. Removing real-data supervision throughout refinement reduces ID success by 8.75 points but OOD success by only 3.75 points. This indicates that failure-targeted simulation corrects observed weaknesses, while real data mainly prevents forgetting. Combining both components, F4R achieves 93.75\% ID and 90.0\% OOD success.

\paragraph{Effects of Co-Training and RL.}
Table~\ref{tab:training_ablation} isolates the contributions of the two policy-refinement stages. Removing Stage I leaves RL to optimize the base policy using simulated interaction and sparse task-success rewards, while retaining auxiliary supervision from the original real-world data. This variant improves the Base by only 13.75 percentage points under ID conditions and 6.25 points under OOD conditions, reaching 76.25\% and 32.5\%, respectively. The limited OOD improvement indicates that the base policy rarely discovers successful behaviors in severe failure regions through sparse-reward exploration alone. In contrast, Stage I co-training without subsequent RL achieves 90.0\% ID and 83.75\% OOD success, showing that failure-conditioned corrective demonstrations provide an effective initialization and account for most of the overall gain. Stage II RL further improves the co-trained policy by 3.75 and 6.25 points, yielding 93.75\% ID and 90.0\% OOD success. These results show that co-training addresses the primary failure modes and alleviates exploration difficulty, while RL further refines closed-loop behaviors beyond fixed demonstrations.

\begin{table}[htbp]
    \centering
    \small
    \renewcommand{\arraystretch}{1.0}

    \begin{adjustbox}{max width=\columnwidth}
        \begin{tabular*}{1\columnwidth}{
            @{\extracolsep{\fill}}
            ccc
            @{}
        }
            \toprule
            Method & ID Avg. & OOD Avg. \\
            \midrule
            Base & 62.5\% & 26.25\% \\
            F4R w/o Stage I SFT & 76.25\% & 32.5\% \\
            F4R w/o Stage II RL & 90\% & 83.75\% \\
            \textbf{F4R} & \textbf{93.75\%} & \textbf{90\%} \\
            \bottomrule
        \end{tabular*}
    \end{adjustbox}
    \caption{\textbf{Ablation of the policy-refinement stages.}
Average real-world success rates (\%) over four tasks under ID and OOD conditions. Stage I denotes failure-driven sim--real co-training, and Stage II denotes reinforcement learning.}
    \label{tab:training_ablation}
\end{table}

\section{Conclusion}

In this work, we present F4R, a failure-driven real-to-sim-to-real framework that transforms deployment failures into targeted training opportunities. F4R employs an agent equipped with a dedicated failure discovery and diagnosis skill to analyze failed trajectories, reconstructs their task-relevant conditions in simulation, and refines the policy through failure-conditioned sim--real co-training followed by targeted RL. Real-world experiments demonstrate consistent improvements under both ID and OOD conditions, outperforming a budget-matched Targeted BC baseline while reducing reliance on additional real-world demonstrations. Multi-round refinement further shows that failures discovered after redeployment can guide subsequent updates, supporting continuous closed-loop improvement. These results establish deployment failures as practical signals for directing simulation generation and policy optimization. Future work will scale F4R to broader manipulation domains and investigate more efficient mechanisms to prioritize and reuse accumulated failure cases as deployment proceeds.

\clearpage
\bibliographystyle{assets/plainnat}
\bibliography{references}
\clearpage
\onecolumn
\setcounter{section}{0}
\setcounter{figure}{0}
\setcounter{table}{0}
\setcounter{equation}{0}
\setcounter{algorithm}{0}
\renewcommand{\thesection}{S\arabic{section}}
\renewcommand{\thesubsection}{\thesection.\arabic{subsection}}
\renewcommand{\thefigure}{S\arabic{figure}}
\renewcommand{\thetable}{S\arabic{table}}
\renewcommand{\theequation}{S\arabic{equation}}
\renewcommand{\thealgorithm}{S\arabic{algorithm}}
\renewcommand{\theHsection}{supp.\arabic{section}}
\renewcommand{\theHfigure}{supp.\arabic{figure}}
\renewcommand{\theHtable}{supp.\arabic{table}}
\renewcommand{\theHequation}{supp.\arabic{equation}}
\renewcommand{\theHalgorithm}{supp.\arabic{algorithm}}
\section*{Supplementary Material}
\addcontentsline{toc}{section}{Supplementary Material}
\section*{Contents}

\noindent\textbf{S1. Failure Recognition and Diagnosis Details}
\dotfill \pageref{supp:sec:diagnosis}\\
\textbf{S2. Failure-Case Reconstruction Details}
\dotfill \pageref{supp:sec:reconstruction}\\
\textbf{S3. Failure-Aware Randomization and Refinement}
\dotfill \pageref{supp:sec:refinement}\\
\textbf{S4. Complete Experimental Setup}
\dotfill \pageref{supp:sec:setup}\\
\textbf{S5. Additional Quantitative Results}
\dotfill \pageref{supp:sec:additional-results}\\
\textbf{S6. Qualitative Results and Limitations}
\dotfill \pageref{supp:sec:qualitative}

\section*{Supplementary Overview}

This supplementary material provides implementation details, complete
experimental protocols, additional quantitative results, qualitative analyses,
and reproducibility information for \FfourR. The material is organized as
follows:
\begin{itemize}[leftmargin=1.4em]
    \item Section~\ref{supp:sec:diagnosis} describes failure recognition and
    diagnosis, including temporal evidence proposal, evidence-board
    construction, prompting, validation, and additional VLM evaluations.
    \item Section~\ref{supp:sec:reconstruction} details manipulation-scene and
    manipulated-object reconstruction and clarifies the human effort required
    by the system.
    \item Section~\ref{supp:sec:refinement} provides failure-aware randomization,
    MimicGen data generation, co-training, and PPO implementation details.
    \item Section~\ref{supp:sec:setup} specifies the robot platform, tasks,
    distributions, success criteria, baselines, and budget matching.
    \item Section~\ref{supp:sec:additional-results} reports complete and additional
    quantitative results.
    \item Section~\ref{supp:sec:qualitative} presents qualitative examples, failure
    cases, limitations, and safety considerations.

\end{itemize}

\FloatBarrier
\section{Failure Recognition and Diagnosis Details}
\label{supp:sec:diagnosis}

\subsection{Input and output}

A task-specific binary success checker first separates successful and failed
deployment rollouts. The diagnostic skill operates only on known failures;
accordingly, its sample-level JSON fixes \texttt{success=no} and focuses on
localizing and attributing the earliest cause. For failed rollout $i$, the
input contains RGB frames with matching indices from a wrist-mounted camera
(\texttt{eyeinhand} in the released code) and a fixed external camera
(\texttt{main}). If the streams have different lengths, only their common
prefix is retained:
\begin{equation}
    \tau_i=\{(I_{i,t}^{w},I_{i,t}^{g})\}_{t=0}^{T_i-1},
    \qquad
    T_i=\min(T_i^{w},T_i^{g}),
    \label{supp:eq:supp_paired_input}
\end{equation}
where $w$ and $g$ denote the wrist and global views. The pipeline generates
temporal proposals, constructs evidence boards, identifies the earliest causal
failure, and aggregates the results into
\begin{equation}
    \mathfrak{D}_i =
    (t_i^\star,o_i^\star,u_i^\star,y_i^\star,
    L_i^\star,E_i^\star,\kappa_i),
    \label{supp:eq:supp_diagnosis}
\end{equation}
where $t_i^\star$ is the earliest visible failure frame, $o_i^\star$ is the
target object, $u_i^\star$ is the first failed operation, $y_i^\star$ is the
failure type, $L_i^\star$ contains one to three scene-factor labels,
$E_i^\star$ is frame-grounded textual evidence, and $\kappa_i$ is confidence.
This diagnosis is incorporated into the downstream failure record
\begin{equation}
    \mathcal{F}_i=(\tau_i,O_i,R_i,z_i),
    \qquad z_i=(u_i^\star,y_i^\star,L_i^\star),
    \label{supp:eq:supp_failure_record}
\end{equation}
where $O_i$ contains relevant objects and $R_i$ contains object--object and
object--container relations. Table~\ref{supp:tab:supp_failure_schema} summarizes
the diagnosis fields and their allowed values.

\begin{table}[!htbp]
    \centering
    \small

    \caption{\textbf{Diagnosis taxonomy and structured output.}}
    \label{supp:tab:supp_failure_schema}
    \setlength{\tabcolsep}{4pt}

    \begin{tabularx}{\linewidth}{@{}p{0.25\linewidth}p{0.27\linewidth}Y@{}}
        \toprule
        Field & Meaning & Allowed values / example \\
        \midrule

        \texttt{first\_\allowbreak{}failure\_\allowbreak{}frame} & Earliest visible anomaly & printed source-frame index \\
        \texttt{first\_\allowbreak{}failure\_\allowbreak{}window} & Supporting evidence board & \texttt{sample\_\allowbreak{}w02} / \texttt{final\_\allowbreak{}state} \\
        \texttt{target\_\allowbreak{}object} & Object manipulated at first failure & task-specific object name \\
        \texttt{operation} & First failed stage & grasp, transport, place, release, uncertain \\
        \texttt{failure\_\allowbreak{}type} & Visible execution outcome & place/grasp failure, outside, slip, tip, collision, occlusion \\
        \texttt{matched\_\allowbreak{}labels} & Failure-relevant scene factors & pose, stiffness, friction, lighting, background, view, clutter \\
        \texttt{critical\_\allowbreak{}events} & Supporting event sequence & window, event type, description \\
        \texttt{evidence} & Auditable textual statement & view and printed frame identifiers \\
        \texttt{confidence} & Evidence consistency & $\kappa_i\in[0.10,0.95]$ \\
        \bottomrule
    \end{tabularx}

\end{table}

\subsection{Temporal evidence proposal}

Processing the full dual-view video as a dense sequence is unnecessarily
expensive and can obscure short interaction events. \FfourR therefore proposes
sparse temporal windows using visual changes in both views. The complete
trajectory is scanned every $s=3$ frames. The in-house videos accompanying this
supplement are recorded at 30 Hz and $1280\times720$, giving an effective scan
rate of 10 Hz. Proposal generation converts each sampled frame to grayscale and
resizes it to $160\times90$ without modifying the source video.

For view $v\in\{w,g\}$, the mean absolute grayscale change is
\begin{equation}
    d_{i,t}^{v}=\frac{1}{HW}\sum_{p=1}^{HW}
    |G_{i,t}^{v}(p)-G_{i,t-s}^{v}(p)|.
    \label{supp:eq:supp_view_change}
\end{equation}
The signal is smoothed with a five-sample moving average $S_5$, robustly
standardized using the trajectory median and median absolute deviation (MAD),
and smoothed again:
\begin{equation}
\begin{aligned}
    a_{i,t}^{v}&=S_5(d_{i,t}^{v}),\\
    \hat d_{i,t}^{v}&=S_5\!\left(
    \frac{a_{i,t}^{v}-\operatorname{med}(a_i^v)}
    {\max(1.4826\,\operatorname{MAD}(a_i^v),10^{-6})}
    \right).
\end{aligned}
\label{supp:eq:supp_robust_change}
\end{equation}
The external view better preserves object--container relations, while the wrist
view is more sensitive to local contact and occlusion. Their scores are fused
as
\begin{equation}
    q_{i,t}=
    \max\!\left(1.2\hat d_{i,t}^{g},
    0.85\hat d_{i,t}^{w}\right).
    \label{supp:eq:supp_event_score}
\end{equation}
For a selected center frame $t_{i,k}$, the corresponding candidate window is
\begin{equation}
    \mathcal{W}_{i,k}
    =
    \left[
    \max(0,t_{i,k}-120),
    \min(T_i-1,t_{i,k}+180)
    \right].
    \label{supp:eq:supp_event_window}
\end{equation}
Only peaks with $q_{i,t}>0$ are eligible. The ten highest-scoring peaks are
retained with a minimum separation of 240 frames. Each peak is expanded by 120
preceding and 180 subsequent frames, retaining additional evidence for
post-contact events such as slipping, bouncing, and incorrect placement. Start
and end context windows of at most 301 frames are always added. Before merging,
the pipeline therefore contains at most 12 intervals. After sorting by start
frame, adjacent intervals are merged when their temporal gap is at most 60
frames. The union defines the merged interval, while its center, score, and
reason are inherited from the stronger proposal. Algorithm~\ref{supp:alg:temporal_proposal}
summarizes the complete procedure, and Table~\ref{supp:tab:supp_proposal_parameters}
lists the parameters used in all experiments.

\begin{algorithm}[t]
    \caption{\textbf{Temporal Evidence Proposal.}
    Sparse windows retain the interaction events and goal-level state
    evidence required for diagnosis.}
    \label{supp:alg:temporal_proposal}

    \begin{algorithmic}[1]
        \Require Synchronized trajectory $\tau_i$
        \Ensure Candidate windows
        $\{\mathcal{W}_{i,k}\}_{k=1}^{K_i}$

        \State Retain the common prefix and scan every third frame.
        \State Resize grayscale frames to $160 \times 90$ and compute
        Eq.~\eqref{supp:eq:supp_view_change}.
        \State Apply five-sample smoothing, MAD standardization, and a second
        smoothing pass.
        \State Fuse the two views using Eq.~\eqref{supp:eq:supp_event_score}.
        \State Retain at most ten positive peaks separated by 240 frames.
        \State Expand each peak by 120 frames before and 180 frames after it.
        \State Add start/end context windows containing at most 301 frames.
        \State Merge intervals separated by at most 60 frames.
        \State Sort the resulting intervals chronologically.
    \end{algorithmic}
\end{algorithm}

\begin{table}[!htbp]
    \centering
    \small
    \caption{\textbf{Temporal-proposal parameters.}}
    \label{supp:tab:supp_proposal_parameters}
    
    \begin{tabularx}{\linewidth}{@{}p{0.29\linewidth}p{0.16\linewidth}Y@{}}
        \toprule
        Parameter & Value & Definition \\
        \midrule

        Scan stride $s$ & 3 frames & Change-estimation interval \\
        Scan resolution & $160\times90$ & Grayscale proposal input \\
        Smoothing $S_5$ & 5 samples & Before and after MAD scaling \\
        Peak threshold & $q_{i,t}>0$ & Positive standardized change \\
        Maximum peaks & 10 & Highest-scoring motion peaks \\
        Pre/post span & 120/180 frames & Peak-window expansion \\
        Minimum peak gap & 240 frames & Peak suppression distance \\
        Merge gap & 60 frames & Maximum gap merged \\
        Context span & $\leq301$ frames & Start/end windows \\
        \bottomrule
    \end{tabularx}
    
\end{table}

\subsection{Dual-view evidence-board construction}

\begin{figure}[!htbp]
    \centering
    \includegraphics[width=1\textwidth]
    {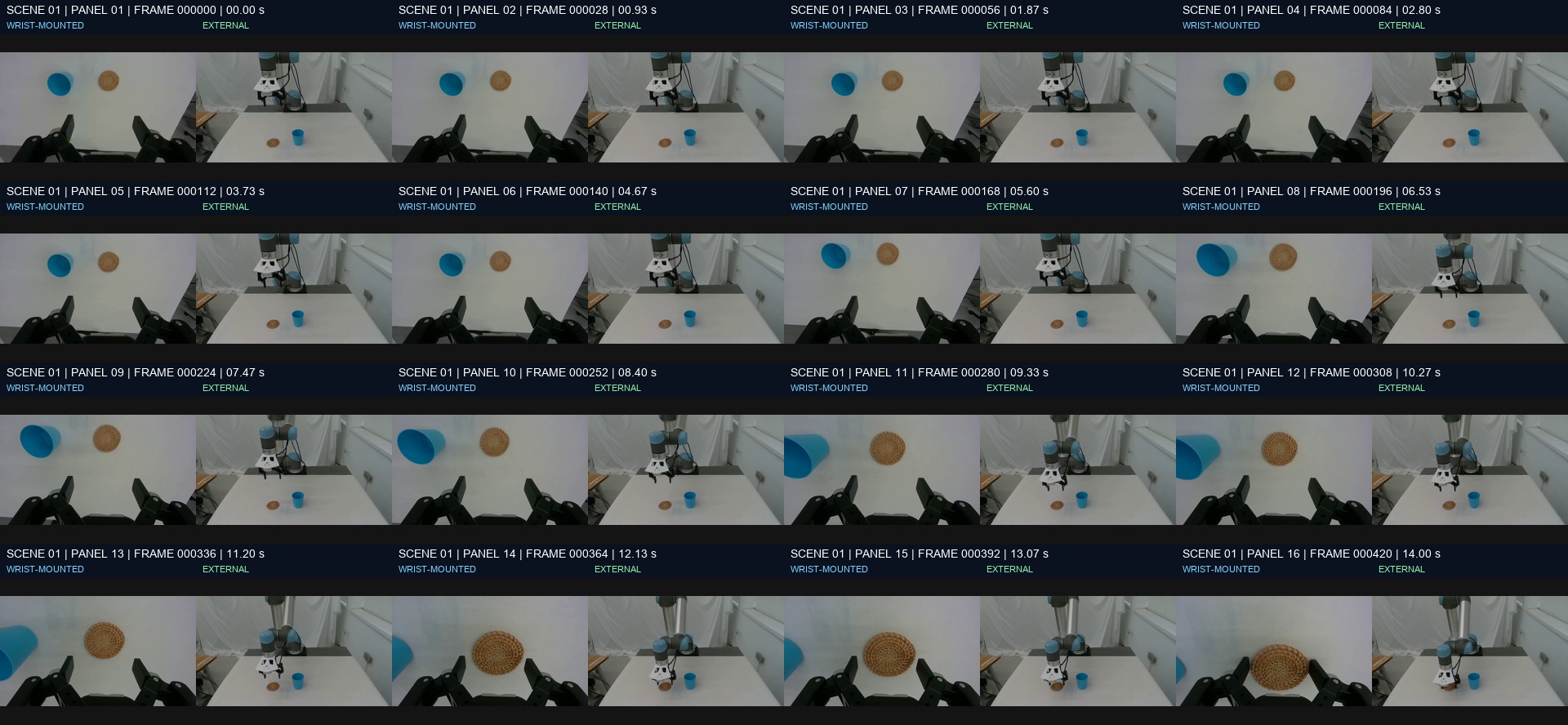}
    \caption{\textbf{Temporal evidence board.} The board contains 16 ordered
    time points in this example. Each panel pairs synchronized wrist and global
    observations. Adaptive sampling becomes denser around the candidate
    interaction while retaining interval-level context. Event boards contain
    up to 24 time points.}
    \label{supp:fig:supp_diagnosis_pipeline}
\end{figure}

Each candidate interval $\mathcal{W}_{i,k}$ is compressed into event board
$\mathcal{B}_{i,k}$. A tile pairs images with the same source-frame index,
placing the wrist view on the left and the external view on the right. Tiles
are ordered chronologically, read from left to right and top to bottom, with
four tiles per row. Each view is resized with aspect-ratio-preserving
letterboxing to $320\times180$ pixels. The paired region is
$640\times180$ with a 42-pixel header containing the sample, window, order, and
original frame index.

An event board contains at most 24 synchronized time points. If interval length
$L\leq24$, all frames are retained. Otherwise, 12 anchors are sampled uniformly
over the interval and 12 dense samples are drawn around the proposal center
with radius $\min(\lfloor L/4\rfloor,8)$. The indices are merged, deduplicated,
sorted, and truncated to 24. The maximum layout is $4\times6$, producing a
$2560\times1332$ JPEG at quality 90.

State board $\mathcal{B}_{i}^{s}$ is independent of motion detection. When
$T_i\leq16$, all synchronized frames are retained; otherwise, the first eight
and final eight are used. Its headers are marked \texttt{START} or \texttt{END}
and have height 46 pixels, yielding a maximum $2560\times904$ JPEG. No
intermediate frames are added: event boards capture interaction dynamics,
whereas the state board tests whether the terminal state satisfies the task.
Each VLM request contains one composite JPEG, corresponding to at most 48
single-view images for an event request and 32 for a state request.
Figure~\ref{supp:fig:supp_diagnosis_pipeline} visualizes a representative temporal
evidence board.

\begin{figure}[!htbp]
    \centering
    \begin{minipage}{0.48\textwidth}
        \centering
        \textbf{Scene 1}\\
        \includegraphics[width=\linewidth]
        {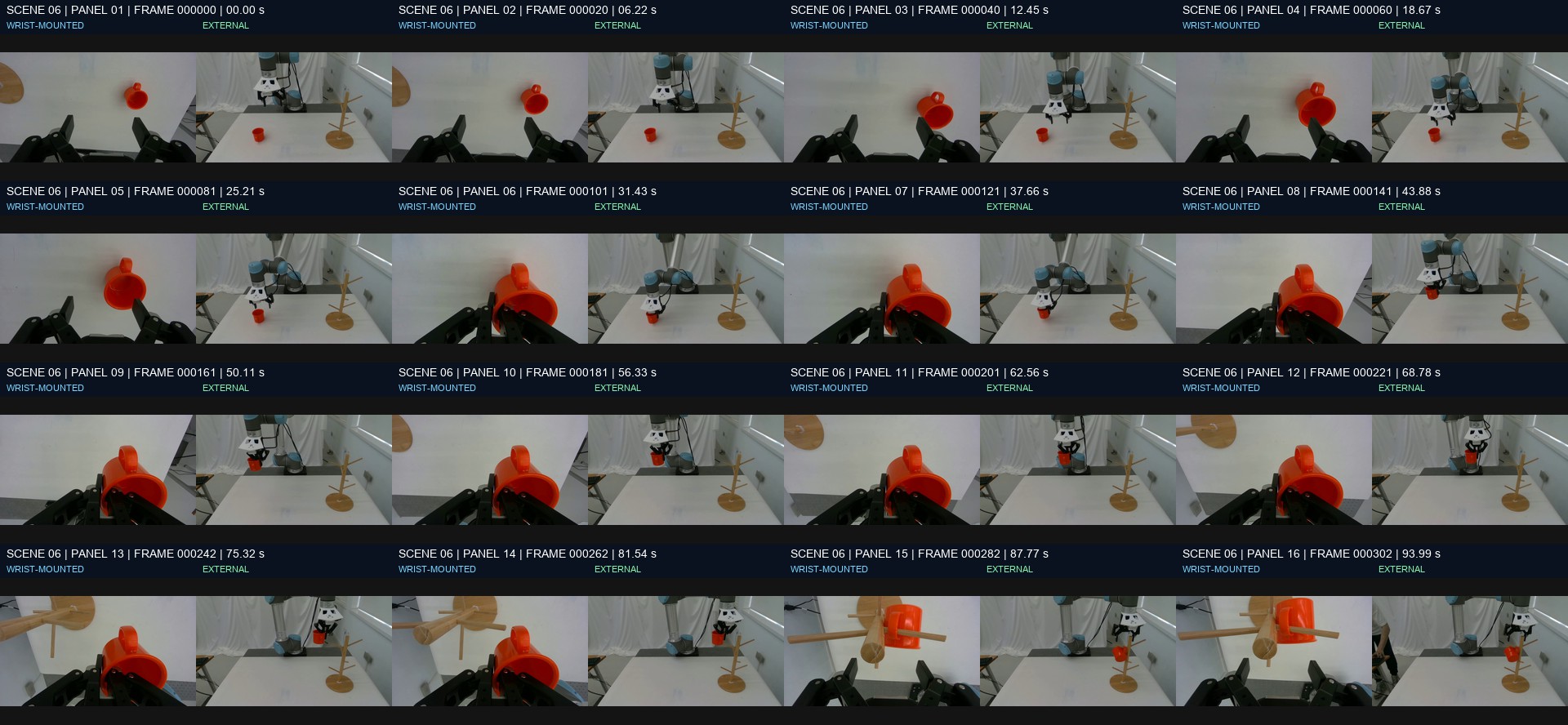}
    \end{minipage}\hfill
    \begin{minipage}{0.48\textwidth}
        \centering
        \textbf{Scene 2}\\
        \includegraphics[width=\linewidth]
        {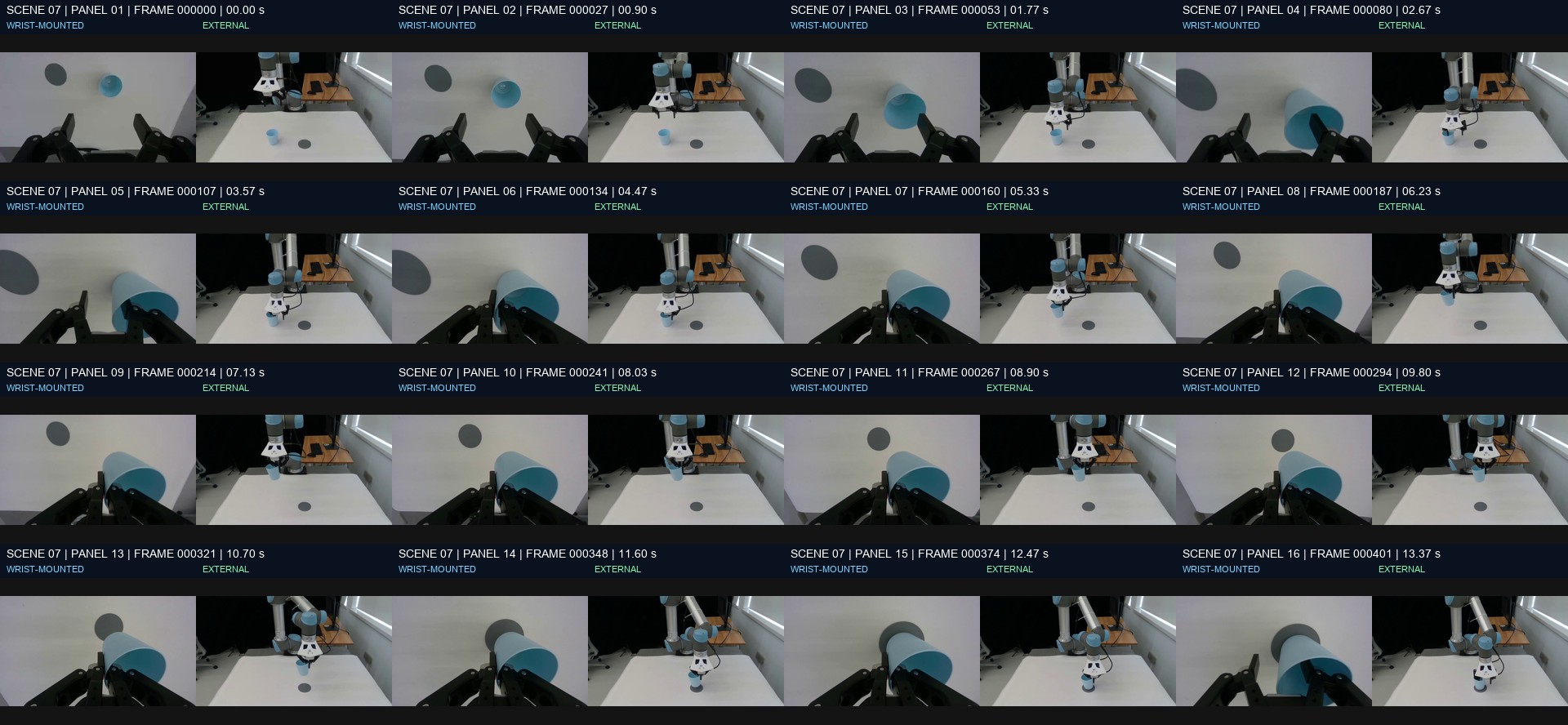}
    \end{minipage}\\
    
    \begin{minipage}{0.48\textwidth}
        \centering
        \textbf{Scene 3}\\
        \includegraphics[width=\linewidth]
        {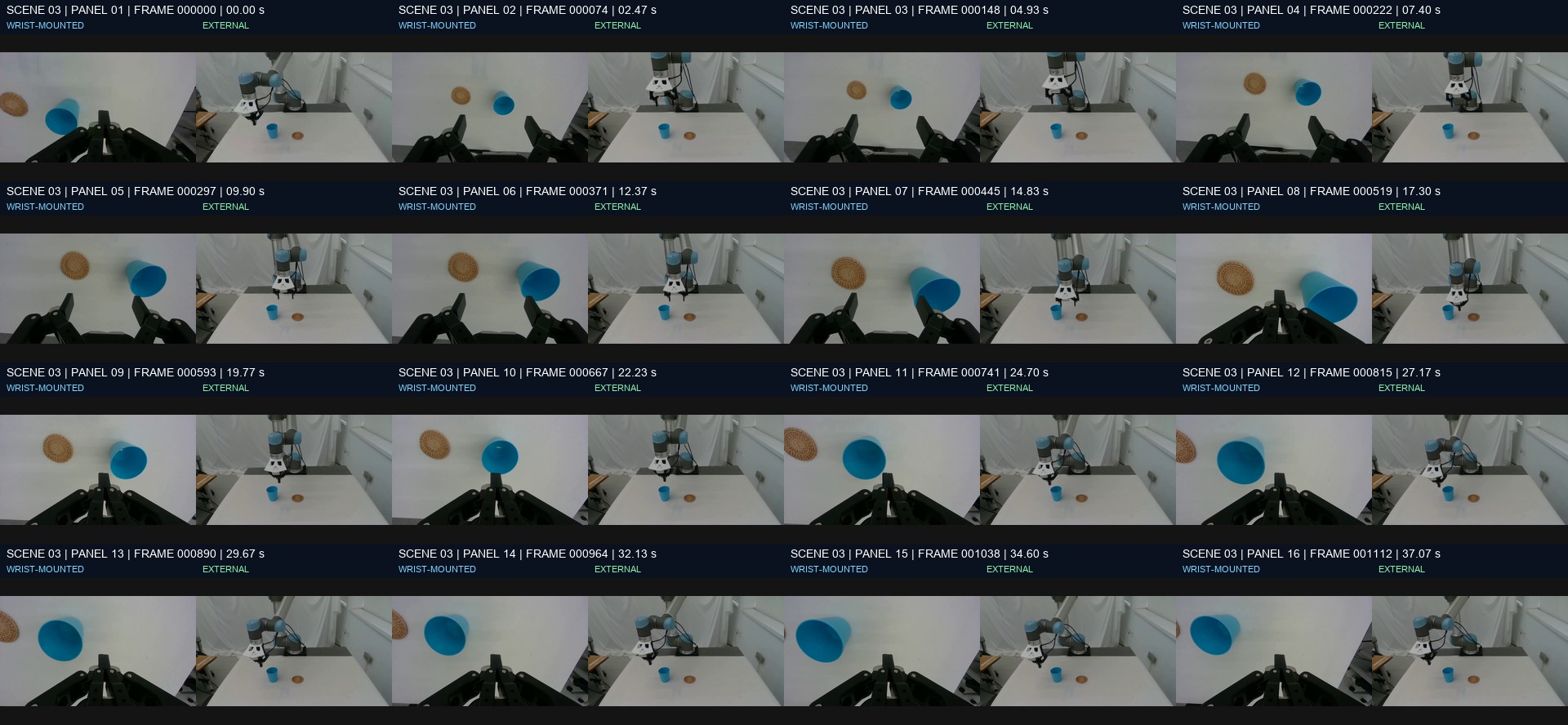}
    \end{minipage}\hfill
    \begin{minipage}{0.48\textwidth}
        \centering
        \textbf{Scene 4}\\
        \includegraphics[width=\linewidth]
        {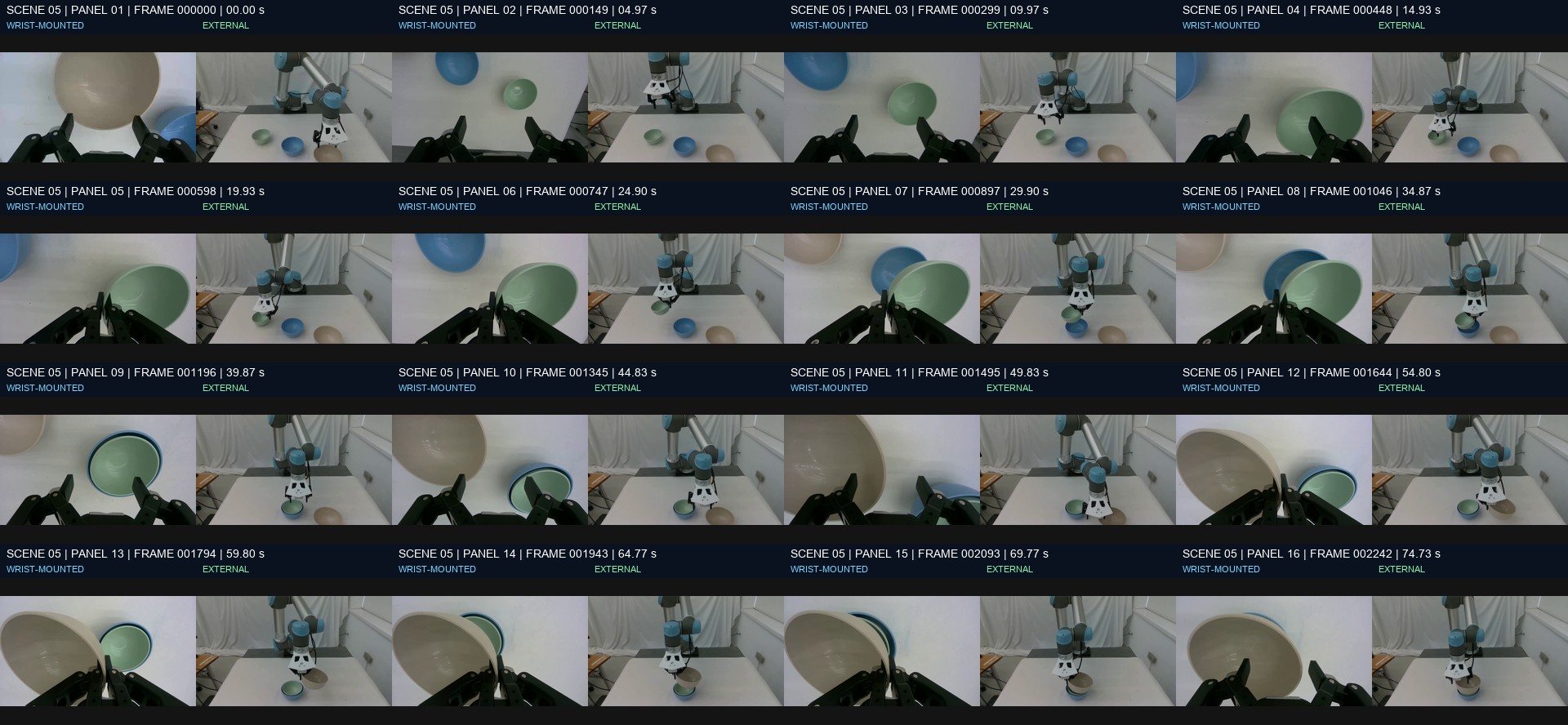}
    \end{minipage}
    \caption{\textbf{Full-rollout storyboards from the in-house dataset.}
    Each board uniformly samples 16 synchronized wrist/external pairs. The
    original frame index and timestamp remain visible in every panel.}
    \label{supp:fig:supp_inhouse_storyboards}
\end{figure}

\paragraph{In-house dual-view rollouts.}
The accompanying qualitative set contains ten videos forming five synchronized
pairs. Odd-numbered videos provide the fixed external view and even-numbered
videos provide the wrist view. Each full-rollout board uniformly samples 16
time points for visual inspection; these overview boards are distinct from the
adaptive event boards sent to the VLM. They use a $4\times4$ layout, place the
wrist view on the left, and report the panel number, source frame, and
timestamp. Exact sampled indices and timestamps are provided in
\url{supplementary_assets/inhouse_storyboards_manifest.json}, together
with the script used to reproduce the boards from paired videos. The five
full-rollout boards are shown in Fig.~\ref{supp:fig:supp_inhouse_storyboards}.

\subsection{Diagnostic prompting}

Diagnosis uses an inference-time controller $\mathcal{A}_{\phi}$ that calls a
pretrained VLM through an OpenAI-compatible chat-completions API. The model,
prompt set, JSON schema, and temperature ($0.05$) are fixed; the controller is
not a separately trained diagnostic network. Given $K_i$ event boards, one
diagnosis uses $K_i+3$ semantic calls: one call for each event board, one
state-board call, one focused call on the earliest positive event board, and
one text-only aggregation call. If no event response contains parseable
positive evidence, the state board is reused as the focused visual input. The
window-level output is
\begin{equation}
    \mathbf r_{i,k}=\mathcal A_{\phi}(\mathcal B_{i,k})
    =(h_{i,k},o_{i,k},u_{i,k},y_{i,k},E_{i,k},c_{i,k}),
    \label{supp:eq:supp_window_diagnosis}
\end{equation}
where $h_{i,k}$ indicates visible failure evidence, $o_{i,k}$ is the target
object, $u_{i,k}$ is the operation, $y_{i,k}$ is the failure type,
$E_{i,k}$ cites observations and printed frame indices, and $c_{i,k}$ is
window-level confidence.

\paragraph{Earliest causal failure.}
Windows are inspected by ascending start frame. The earliest window with
\texttt{contains\_\allowbreak{}failure\_\allowbreak{}evidence=true} and parseable JSON becomes the focused
board. The focused pass scans from its first tile and returns the first printed
frame at which a visible anomaly can explain eventual failure. Later drops,
empty transport motions, and incomplete terminal states are treated as
consequences when an earlier cause is visible. If the first tile already
contains an abnormal state, the system reports only its earliest visible time
and lowers confidence rather than treating that frame as the physical onset.

\paragraph{Prompt templates.}
All visual calls share the system instruction:
\begin{quote}\small
\texttt{You analyze robot manipulation evidence and return strict JSON only.}
\end{quote}
The event prompt states that the sample is a known failed trial, specifies the
left/right view convention and frame headers, enumerates the allowed operations
and failure types, and requests
\texttt{contains\_\allowbreak{}failure\_\allowbreak{}evidence}, \texttt{failure\_\allowbreak{}frame\_\allowbreak{}range},
\texttt{matched\_\allowbreak{}labels}, visible objects, target object, description,
frame-grounded evidence, and confidence. The state prompt compares only
\texttt{START} and \texttt{END} tiles and returns objects present initially,
inside the container, still outside, or missing at termination. The focused
prompt requests the first visible anomaly, copying
\texttt{first\_\allowbreak{}failure\_\allowbreak{}frame} from the printed header rather than the panel
index. Finally, the text-only aggregation prompt combines window, state, and
focused JSON; it uses the focused result for the first failed subtask and
repeated event/state evidence to verify failure type and scene factors.
Table~\ref{supp:tab:supp_diagnostic_calls} summarizes the inputs and structured
outputs of the four calls.

\begin{table}[!htbp]
    \centering
    \small
    \caption{\textbf{Diagnostic calls and structured outputs.} Each visual
    request contains one composite JPEG and returns strict JSON.}
    \label{supp:tab:supp_diagnostic_calls}
    \setlength{\tabcolsep}{4pt}
    \begin{tabularx}{\linewidth}{@{}p{0.17\linewidth}p{0.34\linewidth}Y@{}}
        \toprule
        Call & Core instruction & Required JSON fields \\
        \midrule

        Event window & Identify visible accident or goal-level failure &
        window, evidence flag, type, frame range, object, operation, labels, evidence, confidence \\
        State board & Compare initial and terminal states; do not infer motion &
        final-state failure, start objects, inside/outside objects, labels, evidence, confidence \\
        Focused re-query & Find earliest anomaly in earliest positive board &
        first frame, target object, operation, type, labels, evidence, confidence \\
        Aggregation & Verify the first cause using all preceding JSON &
        first failed subtask, critical events, labels, evidence, sample confidence \\
        \bottomrule
    \end{tabularx}
    
\end{table}

The released v2 script does not inject a separate free-form language
instruction into these four prompts. Instead, task semantics are supplied by
the configured candidate-label list, object aliases, and the preceding binary
success checker. Applying the analyzer to a new task therefore requires
replacing these task-specific entries, while the temporal proposal, board
construction, call order, and JSON interface remain unchanged.

\paragraph{Taxonomy and confidence.}
Allowed operations are \texttt{grasp\_\allowbreak{}from\_\allowbreak{}table},
\texttt{transport\_\allowbreak{}to\_\allowbreak{}bowl}, \texttt{place\_\allowbreak{}into\_\allowbreak{}bowl}, \texttt{release}, and
\texttt{uncertain}. Failure types are \texttt{place\_\allowbreak{}failed},
\texttt{grasp\_\allowbreak{}failed}, \texttt{object\_\allowbreak{}left\_\allowbreak{}outside},
\texttt{object\_\allowbreak{}slipped}, \texttt{basket\_\allowbreak{}tipped}, \texttt{collision},
\texttt{occlusion}, and \texttt{uncertain}. Scene factors include pose, contact
stiffness, friction, illumination, background, camera viewpoint, and tabletop
clutter; task-specific labels may replace this default set. The aggregator
self-reports $\kappa_i\in[0.10,0.95]$ from evidence consistency. This score is
not a calibrated probability and is not thresholded directly.

For the multi-object collection implementation, the candidate task labels are
failure to place an object in the bowl, grasp-time slipping, collision with a
neighboring object, grasping the wrong object, insertion failure, and target
occlusion. Appearance aliases map recognizable objects to specific fruit names
(mangosteen, banana, green apple, and red apple), so the failed subtask records
the manipulated object rather than only its color and shape. Other tasks can
replace both the aliases and candidate labels without changing the inference
procedure.

For ViFailback, we retain the benchmark's native cause labels---task planning,
gripper 6D pose, gripper state, and human intervention. They are used only in
the public-benchmark adapter and are not merged with the F4R failure-type or
scene-factor taxonomies.

\paragraph{Focused pass, caching, and transport retries.}
The focused pass is part of every diagnosis rather than a stochastic
resampling step. After the event calls screen all boards, only the earliest
positive board is resubmitted with a narrower request for failure frame,
target object, operation, and failure type. It does not add images, reorder the
board, request reflection, or resample the interval. Per-window results, the
state result, and the focused result are cached separately; \texttt{--resume}
skips a completed sample, while an interrupted run reuses existing
intermediate JSON. API transport or timeout errors repeat the identical
request up to four times with waits of 5, 10, 15, and 20 seconds. These retries
do not change the evidence or count as new diagnosis attempts.

\paragraph{Simulation-side validation and fallback.}
For diagnosis--reconstruction attempt $a$, the initial policy is evaluated in
the reconstructed distribution $p_F^{(a)}(\mathcal M)$, yielding success
$s^{(a)}$. If $s^{(a)}\leq0.70$, the environment reproduces the observed
weakness and $\mathcal F_i^{(a)}$ is accepted. If $s^{(a)}>0.70$, diagnosis and
reconstruction are repeated. After three consecutive rejections, \FfourR
switches to broad domain randomization. Visual diagnosis and caching are
handled by the analyzer, while this validation loop is executed by the
upper-level \FfourR workflow.

\paragraph{Diagnosis-to-randomization export.}
The released exporter converts the accepted diagnosis into an initial set of
randomization targets, summarized in
Table~\ref{supp:tab:supp_diagnosis_to_dr}. Multiple supported labels are merged. The
reset window begins before $t_i^\star$ by
$\operatorname{clip}(0.05t_i^\star,30,180)$ frames; when no valid frame is
available, the fallback margin is 90 frames. This export provides an auditable
initial configuration that the upper-level workflow further constrains using
task feasibility and reconstruction metadata.

\begin{table}[!htbp]
    \centering
    \small
    \caption{\textbf{Diagnosis-to-randomization mapping implemented by the
    diagnostic skill.}}
    \label{supp:tab:supp_diagnosis_to_dr}
    \setlength{\tabcolsep}{4pt}
    \begin{tabularx}{\linewidth}{@{}p{0.33\linewidth}Y@{}}
        \toprule
        Diagnosed condition & Exported randomization targets \\
        \midrule

        Object not placed in bowl & bowl position, opening orientation, object initial pose \\
        Object slips during grasp & friction, object mass, contact parameters, gripper closing speed \\
        Collision with nearby object & clutter density, obstacle position \\
        Wrong object grasped & similar objects, color, texture, occlusion \\
        Insertion failure & hole-position offset, peg angle, clearance, friction \\
        Target occluded & camera viewpoint, illumination, occluder placement \\
        \bottomrule
    \end{tabularx}
    
\end{table}

\paragraph{Reproducible execution.}
The implementation is divided into deterministic preprocessing and cached VLM
inference. The released commands first detect windows with stride 3, top-10
peaks, 120/180-frame context, and 240-frame suppression; then generate event
and state boards; run GPT-5.5 inference; and export review and randomization
records. The associated scripts and their default arguments are included in
\url{supplementary_assets/robot-video-failure-analyzer}. Intermediate outputs
are stored as \texttt{event\_\allowbreak{}windows.json}, per-window storyboard manifests,
cached JSON responses, a sample-level summary, a review CSV, and
\texttt{failure\_\allowbreak{}conditioned\_\allowbreak{}dr\_\allowbreak{}v2.json}. Responses are parsed as strict
JSON; if surrounding text is returned, the first complete JSON object is
recovered. Unparseable event responses are marked and cannot become the
focused positive window. Returned labels are filtered against the configured
candidate list before export.

The expected directory layout is
\texttt{sample\_\allowbreak{}id/eyeinhand/*.jpg} and
\texttt{sample\_\allowbreak{}id/main/*.jpg}; filenames are naturally sorted before the
common prefix is selected. The inference script reads an OpenAI-compatible
endpoint from \texttt{AIO\_\allowbreak{}BASE\_\allowbreak{}URL} and its key from
\texttt{AIO\_\allowbreak{}API\_\allowbreak{}KEY}. Table~\ref{supp:tab:supp_diagnosis_files} lists the exact
stage-to-artifact interface.

\begin{table}[!htbp]
    \centering
    \small
    \caption{\textbf{Diagnostic-skill scripts and generated artifacts.}}
    \label{supp:tab:supp_diagnosis_files}
    \setlength{\tabcolsep}{3.5pt}
    \begin{tabularx}{\linewidth}{@{}p{0.38\linewidth}YY@{}}
        \toprule
        Script & Function & Primary output \\
        \midrule

        \texttt{detect\_\allowbreak{}event\_\allowbreak{}windows.py} & Dual-view change scanning and window merging & \texttt{event\_\allowbreak{}windows.json/.csv} \\
        \texttt{make\_\allowbreak{}event\_\allowbreak{}storyboards.py} & Adaptive event-board construction & event JPEGs and manifest \\
        \texttt{make\_\allowbreak{}state\_\allowbreak{}comparisons.py} & First-eight/last-eight state-board construction & state JPEGs and manifest \\
        \texttt{analyze\_\allowbreak{}failure\_\allowbreak{}v2.py} & Window, state, focused, and aggregate VLM calls & cached and sample-level JSON \\
        \texttt{export\_\allowbreak{}review\_\allowbreak{}csv.py} & Human-readable diagnostic audit & \texttt{review\_\allowbreak{}v2.csv} \\
        \texttt{export\_\allowbreak{}failure\_\allowbreak{}conditioned\_\allowbreak{}dr.py} & Failure-aware reset and randomization export & \texttt{failure\_\allowbreak{}conditioned\_\allowbreak{}dr\_\allowbreak{}v2.json} \\
        \bottomrule
    \end{tabularx}
    
\end{table}

\begin{figure}[!htbp]
    \centering
    \begin{minipage}{0.32\textwidth}
        \centering
        \textbf{Correct: Coke can}\\
        \includegraphics[width=\linewidth]
        {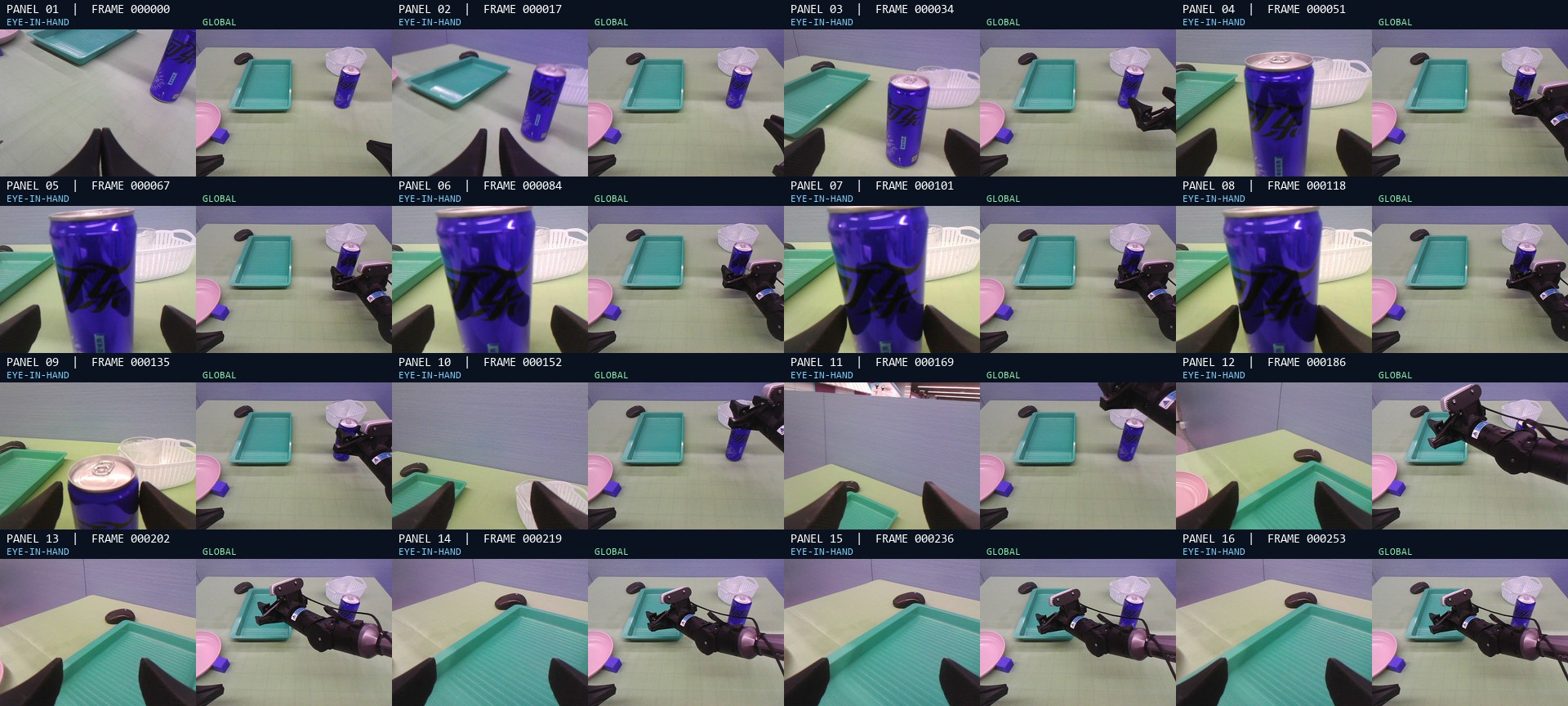}
    \end{minipage}\hfill
    \begin{minipage}{0.32\textwidth}
        \centering
        \textbf{Correct: Spatula}\\
        \includegraphics[width=\linewidth]
        {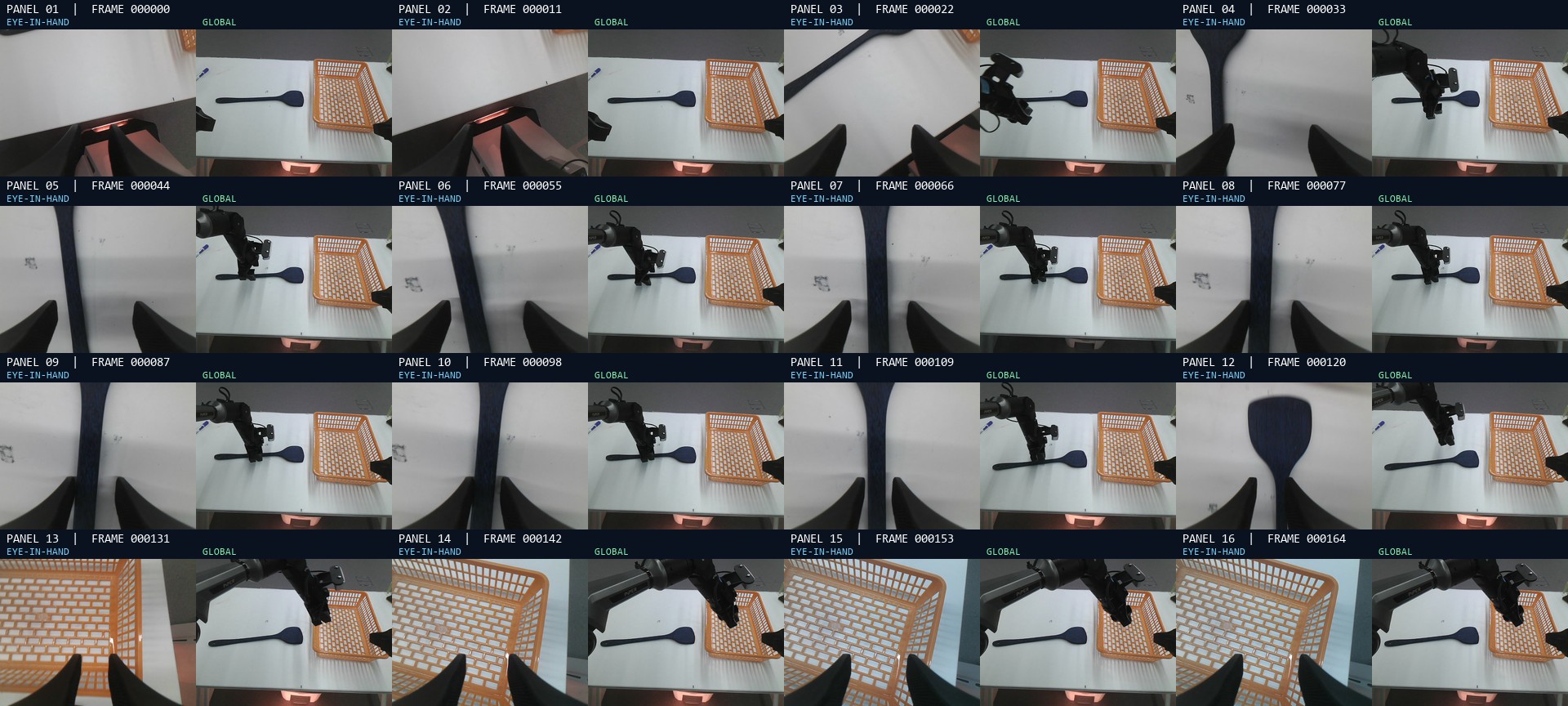}
    \end{minipage}\hfill
    \begin{minipage}{0.32\textwidth}
        \centering
        \textbf{Correct: Cube}\\
        \includegraphics[width=\linewidth]
        {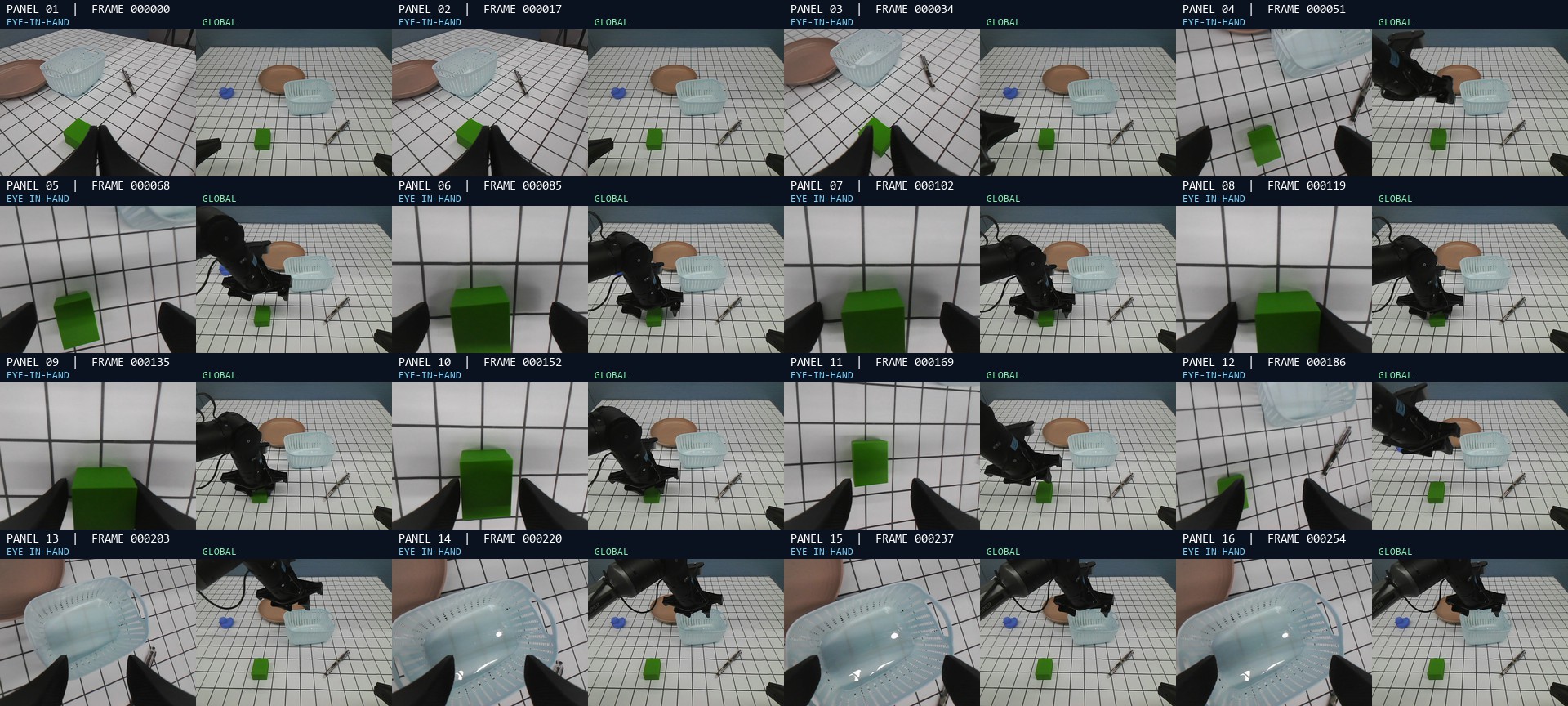}
    \end{minipage}\\[5pt]
    \begin{minipage}{0.48\textwidth}
        \centering
        \textbf{Incorrect: Duck}\\
        \includegraphics[width=\linewidth]
        {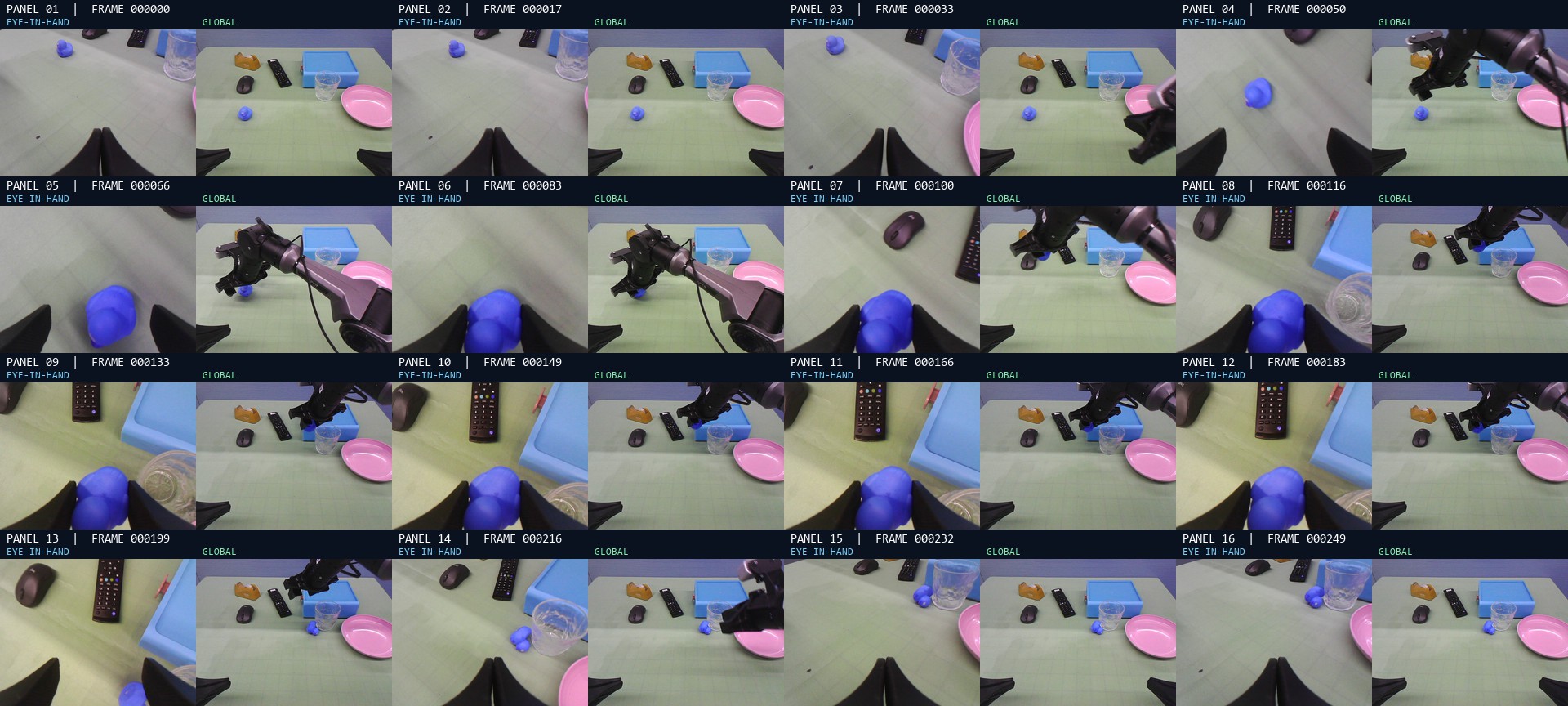}
    \end{minipage}\hfill
    \begin{minipage}{0.48\textwidth}
        \centering
        \textbf{Incorrect: Marker}\\
        \includegraphics[width=\linewidth]
        {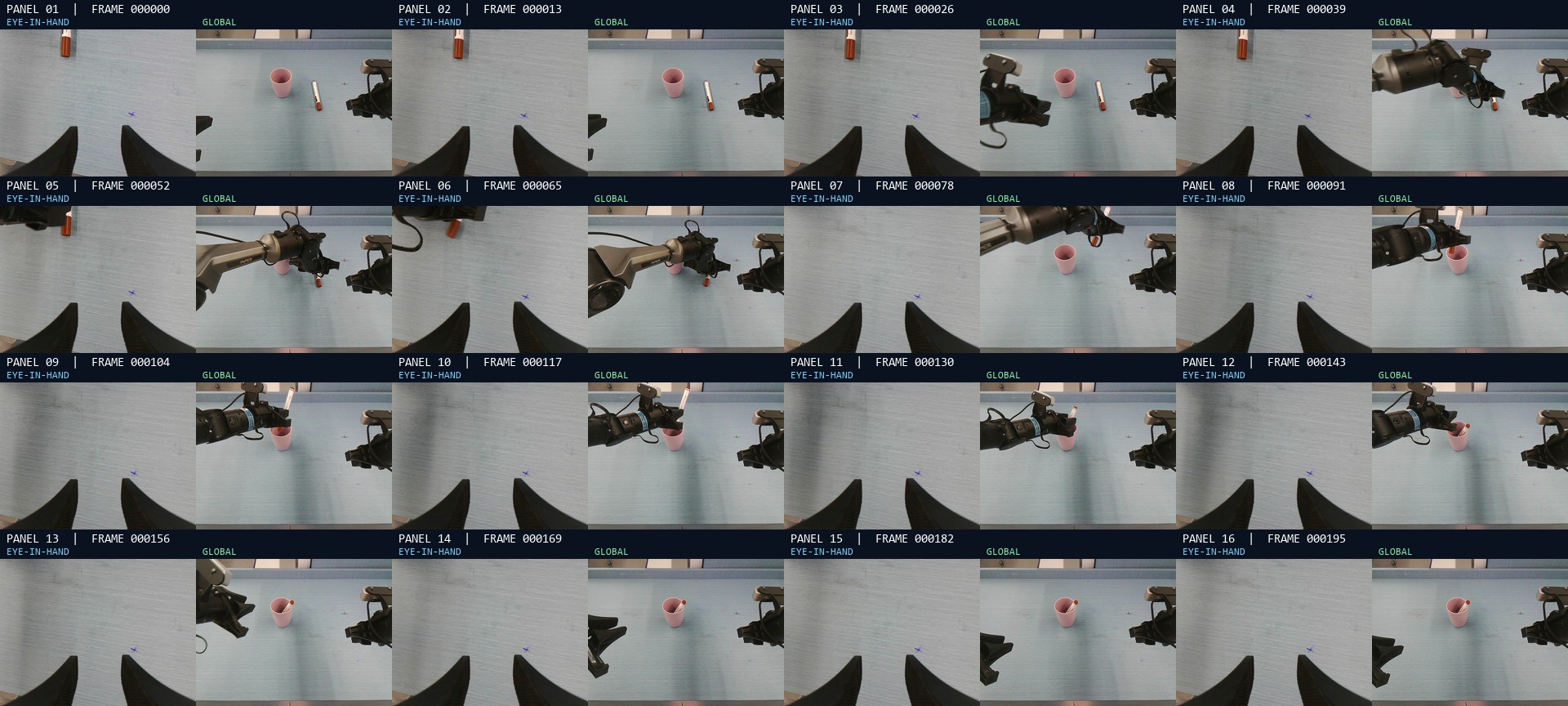}
    \end{minipage}
    \caption{\textbf{Representative diagnosis cases.} The first row shows
    three correctly diagnosed rollouts, while the second row shows two
    incorrect ViFailback cases. A diagnosis is correct only when both cause and
    failed subtask match the annotation. The error cases substitute a salient
    downstream consequence for an earlier causal failure.}
    \label{supp:fig:supp_diagnosis_examples}
\end{figure}

\paragraph{Public-benchmark diagnostic audit.}
Table~\ref{supp:tab:supp_diagnostic_audit} reports the five cases visualized in
Fig.~\ref{supp:fig:supp_diagnosis_examples}. Frame localization is reported
separately from cause/subtask accuracy. The three correct cases identify the
right cause and stage but localize it late, whereas the two errors confuse an
obvious terminal consequence with an earlier failure.

\begin{table}[!htbp]
    \centering
    \small
    \caption{\textbf{Qualitative ViFailback audit.} Prediction/annotation pairs
    for the five visualized cases.}
    \label{supp:tab:supp_diagnostic_audit}
    \setlength{\tabcolsep}{3.5pt}
    \begin{tabularx}{\linewidth}{@{}p{0.08\linewidth}p{0.11\linewidth}p{0.29\linewidth}p{0.12\linewidth}Y@{}}
        \toprule
        Result & Sample & Prediction / annotation & Frame (pred./GT) & Evidence or error \\
        \midrule

        Correct & Coke can & gripper state, subtask 2 / same & 152 / 112 &
        Can remains on table after gripper withdrawal \\
        Correct & Spatula & gripper state, subtask 2 / same & 120 / 75 &
        Arm transports while spatula remains on table \\
        Correct & Green cube & gripper state, subtask 2 / same & 203 / 137 &
        Transport continues after cube is lost \\
        Error & Duck & task planning, subtask 1 / 6D pose, subtask 3 & 50 / 175 &
        Similar object masks later pose error near cup \\
        Error & Marker & 6D pose, subtask 3 / task planning, subtask 1 & 130 / 63 &
        Rim placement is mistaken for the first cause \\
        \bottomrule
    \end{tabularx}
    
\end{table}

\subsection{Autonomous diagnosis benchmark}

We evaluate the diagnosis module on 80 rollout trajectories from eight
tabletop tasks, with ten trajectories per task. Human annotators label both
the earliest failure stage and its underlying cause. A prediction is counted
as correct only when both labels match the annotation. \FfourR uses GPT-5.5
with our robot-video-failure-analyzer skill for this experiment.

As shown in Table~\ref{supp:tab:supp_diagnosis_accuracy}, 69 of 80 trajectories are
correctly diagnosed on the first attempt, yielding $86.25\%$ accuracy. A second
attempt resolves five additional cases and increases cumulative accuracy to
$92.5\%$. Of the six unresolved trajectories, four involve subtle camera
viewpoint shifts and two involve illumination changes. These cases require
fine-grained comparisons with earlier frames to distinguish visual changes
from task events, revealing a remaining limitation in temporal evidence
retention and reasoning.

\begin{table}[!htbp]
    \centering
    \small
    \caption{\textbf{Autonomous failure diagnosis.} Accuracy after one and two
    attempts on 80 trajectories across eight tasks.}
    \label{supp:tab:supp_diagnosis_accuracy}
    \setlength{\tabcolsep}{5pt}
    \begin{tabular}{lcc}
        \toprule
        Outcome & Count & Percentage \\
        \midrule
        Correct after one attempt & 69/80 & 86.25\% \\
        Correct within two attempts & 74/80 & 92.50\% \\
        Remaining: camera viewpoint & 4/80 & 5.00\% \\
        Remaining: illumination & 2/80 & 2.50\% \\
        \bottomrule
    \end{tabular}
\end{table}

\paragraph{Annotation protocol.}
All 80 inputs are known failed trajectories. Each trajectory is manually
annotated with the earliest failed operation and its underlying cause. A
prediction is counted as correct only when both fields match the annotation;
later consequences are not accepted as substitutes for the earliest cause.

\subsection{Complete ViFailback evaluation}

\begin{figure}[!htbp]
    \centering
    \includegraphics[width=0.9\textwidth]
    {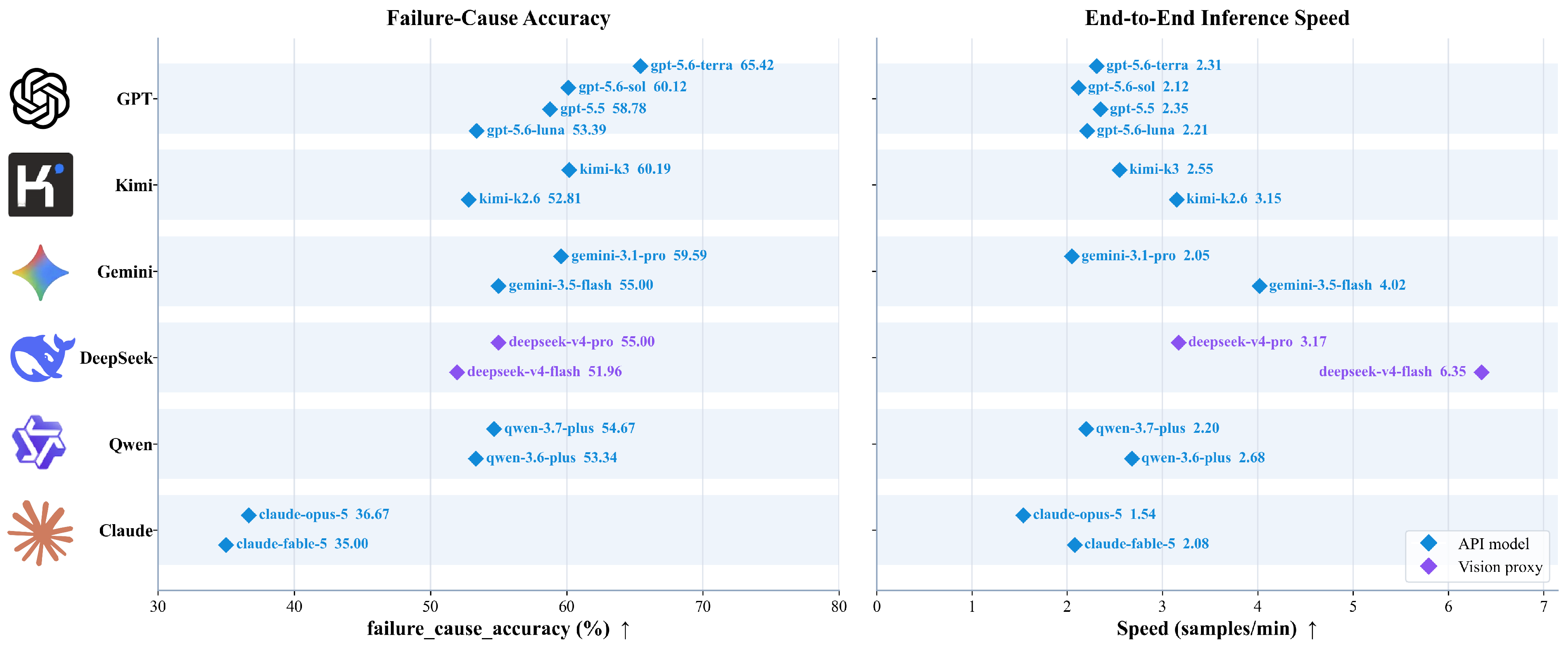}
    \caption{\textbf{Complete ViFailback results.} Failure-cause accuracy and
    end-to-end inference speed of mainstream VLM configurations without
    benchmark-specific adaptation.}
    \label{supp:fig:supp_vifailback_results}
\end{figure}

We additionally evaluate multiple mainstream VLM backends on the public
ViFailback benchmark~\citep{zeng2025diagnose}. All models receive the same
storyboard input and prompt without benchmark-specific adaptation. This
evaluation is separate from our 80-trajectory benchmark: ViFailback measures
general failure attribution on a public dataset, whereas our benchmark uses
task-specific dual-view rollouts and requires the joint prediction of failure
stage and cause. Figure~\ref{supp:fig:supp_vifailback_results} reports the complete
backend comparison.

GPT-5.6-Terra achieves the highest failure-cause accuracy ($65.42\%$), while
Gemini-3.5-Flash provides the highest throughput (4.02 samples/min). Claude
models frequently collapse predictions into the gripper 6D-pose category,
suggesting difficulty separating transient gripper-state and task-planning
failures from pose errors. More importantly, even the best backend remains
below $70\%$ without task-specific adaptation. This result motivates the
simulation-side validation and broad-randomization fallback used by \FfourR,
rather than treating a single VLM diagnosis as ground truth.

\FloatBarrier
\section{Failure-Case Reconstruction Details}
\label{supp:sec:reconstruction}

\subsection{Manipulation-scene reconstruction}

\begin{figure}[!htbp]
    \includegraphics[width=\textwidth]{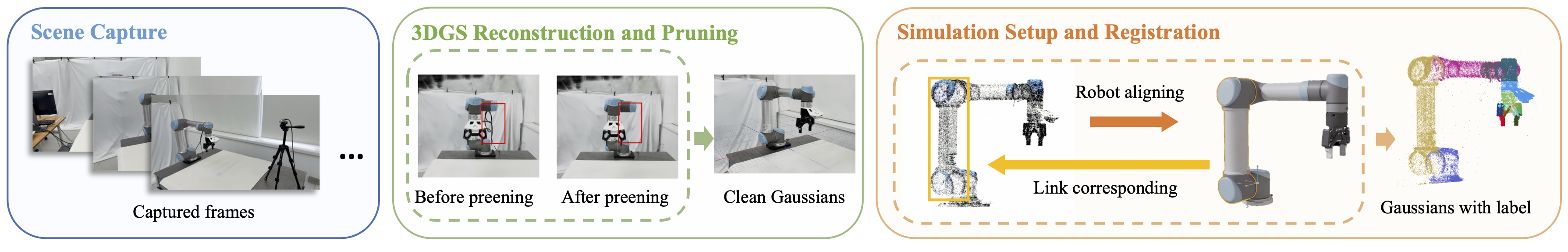}
    \caption{\textbf{Manipulation-scene reconstruction.} The scene is first reconstructed into a 3D Gaussian representation from captured videos. After pruning reconstruction outliers, the Gaussian primitives are aligned with the simulated robot using a scale-aware affine transformation. The aligned robot mesh is then used to assign link-level labels to the Gaussians, enabling the simulation-driven motion of Gaussian representations.}
    \label{supp:fig:supp_scene_pipeline}
\end{figure}

To enable photorealistic simulation rendering while minimizing the manual effort required for environment reconstruction, we propose a Gaussian-based rendering and simulation coupling pipeline. Specifically, we employ 3D Gaussian Splatting (3DGS)~\citep{kerbl3Dgaussians} as the rendering frontend and Isaac Sim as the simulation backend. The robot appearance is represented by link-level Gaussian primitives, which are associated with the corresponding robot links and driven by the rigid transformations provided by the simulator. Fig.~\ref{supp:fig:supp_scene_pipeline} illustrates the overall reconstruction and simulation integration workflow.

We first capture the target manipulation scene using an iPhone camera. During acquisition, the camera intrinsics remain fixed by locking the focal length and avoiding camera switching. The captured video is reconstructed into a photorealistic Gaussian representation using 3DGS. Since the reconstruction may contain irrelevant artifacts, such as floating Gaussians and robot cables, we perform a lightweight cleanup procedure to remove these components and isolate the robot-related Gaussians for subsequent alignment.

Meanwhile, we import the robot model into Isaac Sim and initialize its configuration to match the real-world robot state. The local-to-world rigid transformations of all robot links are extracted from Isaac Sim and used to place the corresponding robot link meshes in the reconstructed scene. To establish geometric correspondence between the reconstructed robot Gaussians primitives and the simulated robot, we adopt a coarse-to-fine registration pipeline consisting of FPFH feature matching, RANSAC-based Sim3 estimation, and Sim3 ICP refinement. The Sim3 transformation compensates for the potential scale discrepancy between the reconstructed Gaussian representation and the simulator-defined robot geometry.

After alignment, the robot Gaussians primitives are associated with individual robot links and transformed into their corresponding local coordinate systems. During simulation, the link-level Gaussian representations are transformed using the local-to-world rigid transformations from Isaac Sim, allowing the rendered robot appearance to faithfully follow the simulated robot motion. For the manipulation scene, the tabletop height is estimated by fitting the tabletop surface from the reconstructed Gaussians and further verified with physical measurements. The detailed reconstruction and simulation parameters are provided in Table~\ref{supp:tab:supp_scene_settings}.

\begin{figure}[!htbp]
    \includegraphics[width=\textwidth]{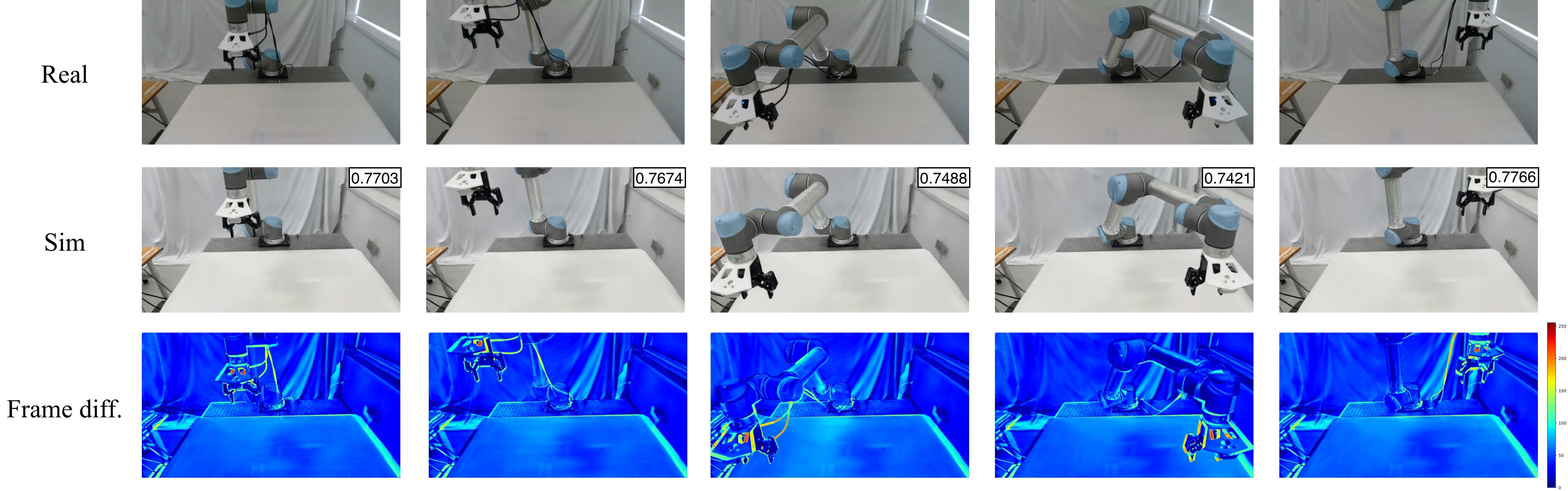}
    \caption{\textbf{Scene reconstruction quality.} The top row shows images captured from the first-person perspective using a real-world RealSense D435i camera; the second row displays images rendered using Gaussian Splatting, with the SSIM between the sim-real images listed in the top-right corner; the third row shows the frame difference.}
    \label{supp:fig:supp_scene_quality}
\end{figure}

With the aligned Gaussian representation and simulation assets, the simulated robot motions and interactions can be faithfully reflected in the rendered observations. This coupling significantly reduces the effort required for constructing simulation environments while maintaining a small visual gap between simulation rendering and real-world observations. Qualitative comparisons between rendered views and real camera observations are presented in Fig.~\ref{supp:fig:supp_scene_quality}.

\begin{table}[!htbp]
    \centering
    \caption{\textbf{Scene reconstruction settings.}}
    \label{supp:tab:supp_scene_settings}
    \begin{tabular}{ll}
        \toprule
        Setting & Value \\
        \midrule
        Smartphone model & IPhone 17 Pro \\
        Capture resolution / frame rate & 1920$\times$1080 / 30 FPS \\
        Capture duration / frame count & $\approx$ 5 min \\
        3DGS optimization iterations & 30000 \\
        3DGS optimization hyperparameters & Default \\
        Scene-to-simulator registration method & FPFH$\rightarrow$Sim3 RANSAC$\rightarrow$Sim3 ICP \\
        Scene setup time & $\approx$ 30 min \\
        Reuse condition & Reused unless the workspace changes \\
        \bottomrule
    \end{tabular}
\end{table}

\subsection{Manipulated-object reconstruction}

\begin{figure}[!htbp]
    \centering
    \includegraphics[width=\linewidth]{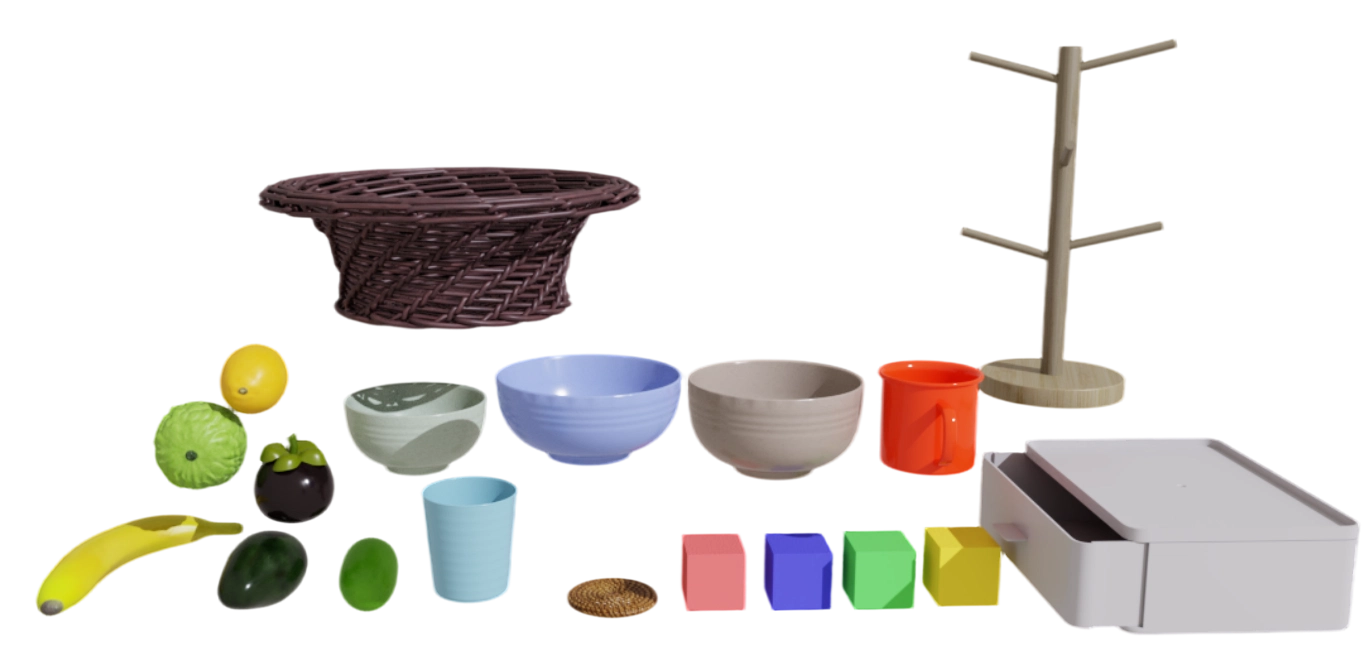}
    \caption{\textbf{Object reconstruction quality.} Rendering of object assets using in the experiments.}
    \label{supp:fig:supp_object_qualitative}
\end{figure}

For object reconstruction, F4R supports two complementary approaches: 
(1) direct generation of simulation-ready assets from images and 
(2) geometric reconstruction from multi-view observations. 
Since image-based asset generation follows existing image-to-3D pipelines, we omit its implementation details and focus on the multi-view reconstruction approach adopted in our experiments. The object assets using in the experiments are shown in Figure~\ref{supp:fig:supp_object_qualitative} and the hyper parameters of object reconstruction are listed in Table~\ref{supp:tab:supp_object_settings}.

Each manipulated object is reconstructed from a COLMAP-style RGB-D dataset~\citep{schoenberger2016sfm} collected by the wrist camera, including calibrated RGB images, metric depth maps, camera intrinsics, poses, and optional object masks. During preprocessing, the camera poses are converted from world-to-camera to base-to-camera transformations. To improve reconstruction accuracy, the depth maps can be refined using a learned depth estimator~\citep{lingbot-depth2026} conditioned on RGB images, raw depth observations, and normalized camera intrinsics. The refined depth is constrained to a valid metric range and combined with object masks before being back-projected into camera-frame point clouds. These point clouds are voxel-downsampled and augmented with surface normals for subsequent registration.

To obtain a consistent object geometry, neighboring observations are associated within a temporal window, and point correspondences are established through nearest-neighbor matching in the world coordinate system based on the initial robot poses. Camera poses are further optimized through a robust nonlinear optimization problem over SE(3), combining point-to-plane geometric alignment with a robot-pose prior that regularizes translation and rotation updates. With the optimized poses, masked TSDF fusion~\citep{curless1996volumetric} produces a metric reconstruction, including a point cloud \(g_i^{\mathrm{pcd}}\) and an initial mesh. The fused geometry is further denoised before asset generation.

The reconstructed geometry is completed into a watertight mesh \(\bar{g}_i^{\mathrm{mesh}}\) using ShapeR~\citep{siddiqui2026shaper}, followed by geometric regularization operations including voxelization, hole filling, iso-surface extraction, smoothing, and optional mesh simplification. For appearance reconstruction, object masks are first propagated from RGB observations using SAM2~\citep{ravi2024sam2}. Informative reference views are then selected according to foreground coverage, image sharpness, and object-view distance, while enforcing angular diversity among selected viewpoints. The selected views are provided to a Hunyuan3D 2.1~\citep{hunyuan3d2025hunyuan3d} texture inpainting workflow implemented in ComfyUI to generate the textured mesh \(g_i^{\mathrm{tex}}\).

The textured mesh is subsequently rendered into a synthetic COLMAP-style dataset containing RGB images, depth maps, object masks, camera parameters, and sampled surface points. These renderings are used to optimize the Gaussian representation~\citep{chen2024pgsr}, producing the final rendering asset \(g_i^{\mathrm{GS}}\).

For articulated objects, an additional semantic and kinematic reasoning module is incorporated following RoboSimGS~\citep{zhao2026high}. Specifically, a multimodal large language model~\citep{bai2025qwen3} can infer movable parts and their kinematic relationships from multi-view renderings, followed by mesh segmentation~\citep{ma2025p3}.

\begin{table}[!htbp]
\centering
\caption{\textbf{Object reconstruction settings.}}
\label{supp:tab:supp_object_settings}
\begin{tabular}{ll}
\toprule
Parameter & Value \\
\midrule
Input depth range & $[0.05, 2.0]$ m \\
Back-projection stride & $1$ px \\
Per-frame voxel size & $0.003$ m \\
Normal radius & $3.0\times$ voxel size \\
Normal max neighbors & $30$ \\
Frame-pair window & $3$ neighboring frames \\
Max correspondence distance & $0.015$ m \\
Max correspondences per pair & $3000$ \\
Pose optimizer & robust least squares, Huber loss \\
Optimizer iterations & $50$ \\
Huber scale & $1.0$ \\
Robot translation prior & $\sigma_t=0.002$ m \\
Robot rotation prior & $\sigma_R=0.2^\circ$ \\
TSDF voxel size & $0.002$ m \\
TSDF truncation distance & $6\times$ voxel size \\
TSDF depth range & $[0.15, 1.20]$ m \\
Output voxel downsampling & $0.0015$ m \\
Statistical outlier removal & $20$ neighbors, std ratio $1.5$ \\
Radius outlier removal & radius $0.005$ m, min neighbors $8$ \\
Mesh generation conditioned images & 16 \\
Mesh generation denoising steps & 25 \\
Mesh inpainting reference images & 4 \\
\bottomrule
\end{tabular}
\end{table}

\subsection{Automation and human Effort}

\begin{table}[!htbp]
    \centering
    \small
    \caption{\textbf{Human-effort breakdown.} Automation boundary of \FfourR.}
    \label{supp:tab:supp_human_effort}
    \setlength{\tabcolsep}{5pt}
    \begin{tabularx}{\linewidth}{@{}p{0.31\linewidth}Yp{0.23\linewidth}p{0.10\linewidth}@{}}
        \toprule
        Stage & Human involvement & Frequency & Typical time \\
        \midrule

        Workspace smartphone capture 
        & One-time manual 
        & New workspace 
        & 3--5 min \\
        
        Scene reconstruction and alignment 
        & Lightweight verification 
        & New/changed workspace 
        & $\sim$20 min \\
        
        Wrist RGB-D object observation 
        & Automatic robot execution 
        & New object/failure case 
        & $\sim$2 min \\
        
        Object asset reconstruction 
        & Automatic 
        & New object 
        & 5--10 min \\
        
        Corrective data generation 
        & Automatic 
        & Each refinement cycle 
        & $\sim$10 min \\
        
        Real-world corrective demonstration 
        & Not required 
        & --- 
        & 0 min \\
        \bottomrule
    \end{tabularx}
\end{table}

The \emph{autonomous closed loop} of \FfourR starts after the initial workspace initialization. The initial smartphone capture and scene setup require one-time human involvement, while subsequent failure detection, reconstruction, simulation refinement, and corrective data generation are automated. Reconstructed scenes and object assets can be reused for unchanged environments. Table~\ref{supp:tab:supp_human_effort} summarizes the automation boundary of each stage.

\FloatBarrier
\section{Failure-Aware Randomization and Policy Refinement}
\label{supp:sec:refinement}

\subsection{Failure-aware domain randomization}

Let $\widehat{\mathcal{M}}_i$ denote the reconstructed simulator environment for
failure $i$, and let $\xi_i^f$ denote its recovered parameters, including object
poses, robot initialization, camera configuration, appearance, and physical
properties. Given failure record $\mathcal{F}_i$, \FfourR samples
\begin{equation}
    \xi\sim p_i^F(\xi\mid\xi_i^f,\mathcal{F}_i),
    \qquad
    \mathcal{M}_{i,\xi}
    =\mathrm{Rand}(\widehat{\mathcal{M}}_i;\xi),
    \label{supp:eq:supp_randomization}
\end{equation}
where $\mathcal{M}_{i,\xi}$ is the randomized instance produced from
$\widehat{\mathcal{M}}_i$. Failure-relevant variables are sampled locally
around the recovered configuration, while unrelated variables are fixed or
weakly perturbed. Samples that violate task semantics, cause initial
collisions, place objects outside the workspace, or make the task infeasible
are rejected.

Across the accepted failures, the training distribution is
\begin{equation}
    p_F(\mathcal{M})=
    \frac{1}{|\mathcal{D}_{\mathrm{fail}}|}
    \sum_{i=1}^{|\mathcal{D}_{\mathrm{fail}}|}
    p_i^F(\mathcal{M}\mid\widehat{\mathcal{M}}_i,\mathcal{F}_i).
    \label{supp:eq:supp_failure_distribution}
\end{equation}
During training, accepted failures are sampled uniformly, so each reconstructed
failure distribution contributes equally without reweighting by frequency,
severity, validation score, or training progress.
Table~\ref{supp:tab:supp_randomization_ranges} specifies the task-conditioned local
variables.

\begin{table}[!htbp]
    \centering
    \small

    \caption{\textbf{Failure-aware randomization ranges.} Active variables and
    sampling distributions for the four refinement tasks.}
    \label{supp:tab:supp_randomization_ranges}
    \setlength{\tabcolsep}{5pt}
    \begin{tabularx}{\linewidth}{@{}p{0.22\linewidth}p{0.29\linewidth}p{0.18\linewidth}Y@{}}
        \toprule
        Task / failure & Variable & Nominal source & Feasibility constraint \\
        \midrule

        Pick Fruits & Fruit pose $(x,y,\psi)$ & diagnosed scene & reachable; no overlap \\
        Pick Fruits & Basket pose / visibility & diagnosed scene & target remains valid \\
        Place Cup on Coaster & Cup pose & diagnosed scene & reachable \\
        Place Cup on Coaster & Coaster pose / wrist visibility & diagnosed scene & valid placement area \\
        Stack Bowls & Bowl poses / placement order & diagnosed scene & stable initial state \\
        Place Block in Drawer & Block pose & diagnosed scene & reachable \\
        Place Block in Drawer & Drawer pose / orientation & diagnosed scene & collision-free; operable \\
        All & Camera extrinsics & calibration & workspace visible \\
        All & Illumination / appearance & recovered scene & physically plausible \\
        All & Friction / mass & asset estimate & stable initialization \\
        \bottomrule
    \end{tabularx}
    
\end{table}

\subsection{Corrective trajectory generation}

\FfourR uses historical successful real-world rollouts as seed trajectories for
MimicGen~\citep{mandlekar2023mimicgen}. MimicGen segments these demonstrations
into object-centric motion segments and adapts them to environments sampled
from $p_F(\mathcal{M})$. If the available successful rollouts do not provide
sufficient grasp coverage, AnyGrasp~\citep{fang2023anygrasp} supplies candidate grasp poses. A
task-specific motion planner or controller then connects the selected grasp to
the remaining manipulation segments. Only trajectories satisfying the
simulation success criterion are retained in
$\mathcal{D}_{\mathrm{sim}}^F$; unsuccessful rollouts are discarded and
resampled for supervised co-training but remain available as interaction
experience during PPO. Table~\ref{supp:tab:supp_mimicgen} reports the resulting data
statistics.

\begin{table}[!htbp]
    \centering
    \small
    \caption{\textbf{Corrective-data generation statistics.}
    Historical seeds denote real-world successful rollouts used to initialize
    MimicGen-style generation. Generated denotes all candidate simulated
    rollouts before filtering. Successful denotes rollouts satisfying the
    simulator success predicate. Final $|\mathcal{D}_{\mathrm{sim}}^F|$
    denotes the number of trajectories retained for supervised co-training.}
    \label{supp:tab:supp_mimicgen}
    \setlength{\tabcolsep}{4pt}
    \begin{tabularx}{\linewidth}{@{}p{0.16\linewidth}YYYYYYY@{}}
        \toprule
        Task & Historical seeds & Generated & Successful & Retained / generated &
        AnyGrasp used & Final $|\mathcal{D}_{\mathrm{sim}}^F|$ & Avg. horizon \\
        \midrule

        Pick Fruits & 24 & 500 & 438 & 82.0\% & 64 / 410 & 410 & 62.5 \\
        Place Cup on Coaster & 22 & 560 & 462 & 76.8\% & 118 / 430 & 430 & 71.3 \\
        Stack Bowls & 28 & 640 & 486 & 70.3\% & 156 / 450 & 450 & 83.7 \\
        Place Block in Drawer & 26 & 620 & 421 & 61.3\% & 142 / 380 & 380 & 96.4 \\
        \bottomrule
    \end{tabularx}
\end{table}

\subsection{Stage I: sim--real co-training}

\FfourR co-trains the failure-targeted simulation trajectories with the
original real-world data:
\begin{equation}
    \mathcal{L}_{\mathrm{CT}}(\theta)
    =
    \mathcal{L}_{\mathrm{SFT}}(\theta;\mathcal{D}_{\mathrm{real}})
    +
    \lambda_{\mathrm{sim}}
    \mathcal{L}_{\mathrm{SFT}}(\theta;\mathcal{D}_{\mathrm{sim}}^F).
    \label{supp:eq:supp_cotraining}
\end{equation}
The simulated data provide failure-specific corrections, while the real data
preserve the policy's existing capabilities and provide a stable
initialization for RL.

\subsection{Stage II: failure-targeted reinforcement learning}

The PPO stage is initialized from the co-trained policy and interacts with 256
parallel environments sampled from $p_F(\mathcal{M})$. \FfourR retains the
original sparse task reward and does not introduce failure-specific reward
shaping. During PPO, an auxiliary SFT loss on the original real-world data
regularizes the policy:
\begin{equation}
    \mathcal{L}_{\mathrm{F4R}}(\theta)
    =
    \mathcal{L}_{\mathrm{PPO}}(\theta;p_F(\mathcal{M}))
    +
    \beta
    \mathcal{L}_{\mathrm{SFT}}(\theta;\mathcal{D}_{\mathrm{real}}).
    \label{supp:eq:supp_ppo}
\end{equation}
Co-training enables the policy to reach meaningful states in severe failure
regions; PPO then improves closed-loop and contact-sensitive behavior. The
complete SFT and PPO configuration is listed in
Table~\ref{supp:tab:supp_training_hyperparameters}.

 \begin{table}[!htbp]
      \centering
      \small
      \caption{\textbf{SFT and PPO hyperparameters.}}
      \label{supp:tab:supp_training_hyperparameters}
      \setlength{\tabcolsep}{5pt}
      \begin{tabularx}{\linewidth}{@{}p{0.15\linewidth}p{0.40\linewidth}Y@{}}
          \toprule
          Stage & Hyperparameter & Value \\
          \midrule

          SFT & Base policy & $\pi_{0.5}$ \\
          SFT & Trainable modules & LoRA\\
          SFT & Optimizer & AdamW \\
          SFT & Learning rate & $2.5\times10^{-5}$ \\
          SFT & GPUs & 4 $\times$ NVIDIA H20 \\
          SFT & Distributed strategy & PyTorch FSDP \\
          SFT & Samples per GPU / global batch & 32 / 128 \\
          SFT & $\lambda_{\mathrm{sim}}$ / real:sim sampling ratio & 1:10 \\
          SFT & Epochs / update steps / warmup & step-based / 30{,}000 / 1{,}000 \\
          SFT & Weight decay / Adam betas / schedule & $10^{-10}$ / $(0.9,0.95)$ / cosine \\
          SFT & Visual resolution / action normalization & 2 views at $224\times224$ / OpenPI norm stats + delta joints \\
          SFT & Action horizon $H$ / control frequency & 50 / 30 Hz \\
          \midrule
          PPO & Parallel environments & 256 \\
          PPO & Update steps & 200 \\
          PPO & Rollout horizon / transitions per update & 900 env steps / 4{,}608 action chunks \\
          PPO & Learning rate / clip ratio & $5\times10^{-6}$ / $[0.9,1.2]$ \\
          PPO & Discount $\gamma$ / GAE $\lambda$ & 0.99 / 0.95 \\
          PPO & Entropy / value coefficients & 0 / 1 \\
          PPO & Epochs per update / minibatch size & 2 / 4{,}608 \\
          PPO & Max gradient norm / advantage normalization & 1.0 / enabled \\
          PPO & KL regularization / early stopping & $\beta_{\mathrm{KL}}=0$ / none \\
          PPO & Auxiliary real-data coefficient $\beta$ & 0.1 \\
          PPO & Trainable modules & Full policy parameters + value head \\
          PPO & Reward & Original sparse task reward \\
          \bottomrule
      \end{tabularx}
  \end{table}

\paragraph{Distributed learner--rollout architecture.}
Four H20 GPUs perform policy inference and optimization, while two RTX 4090 GPUs run Isaac Lab 3.0 and collect rollouts. The nodes form a Ray cluster over SSH. Training is synchronous: each PPO iteration synchronizes the actor parameters, collects a complete rollout batch, and then updates the policy. Parameters are broadcast every iteration (\texttt{weight\_\allowbreak{}sync\_\allowbreak{}interval}=1) and before evaluation. The scheduler and Arena subprocess queues are unbounded (\texttt{maxsize}=0). For \textit{Stack Bowls}, each update collects $256\times900=230{,}400$ low-level transitions, equivalent to 4,608 action chunks. Throughput was not directly logged and can be derived as $115{,}200/T_{\mathrm{step}}$ environment steps/s or $4{,}608/T_{\mathrm{step}}$ action chunks/s.

\FloatBarrier
\section{Complete Experimental Setup}
\label{supp:sec:setup}

\subsection{Robot platform and calibration}

The real-world platform consists of a 6-DoF UR5 robot and a Robotiq 2F-85
gripper. Two Intel RealSense D435i cameras provide visual observations: a fixed
third-person camera captures global context, and a wrist-mounted camera
captures fine-grained interaction details.

\begin{figure}[!htbp]
    \centering
    \includegraphics[width=0.65\linewidth]{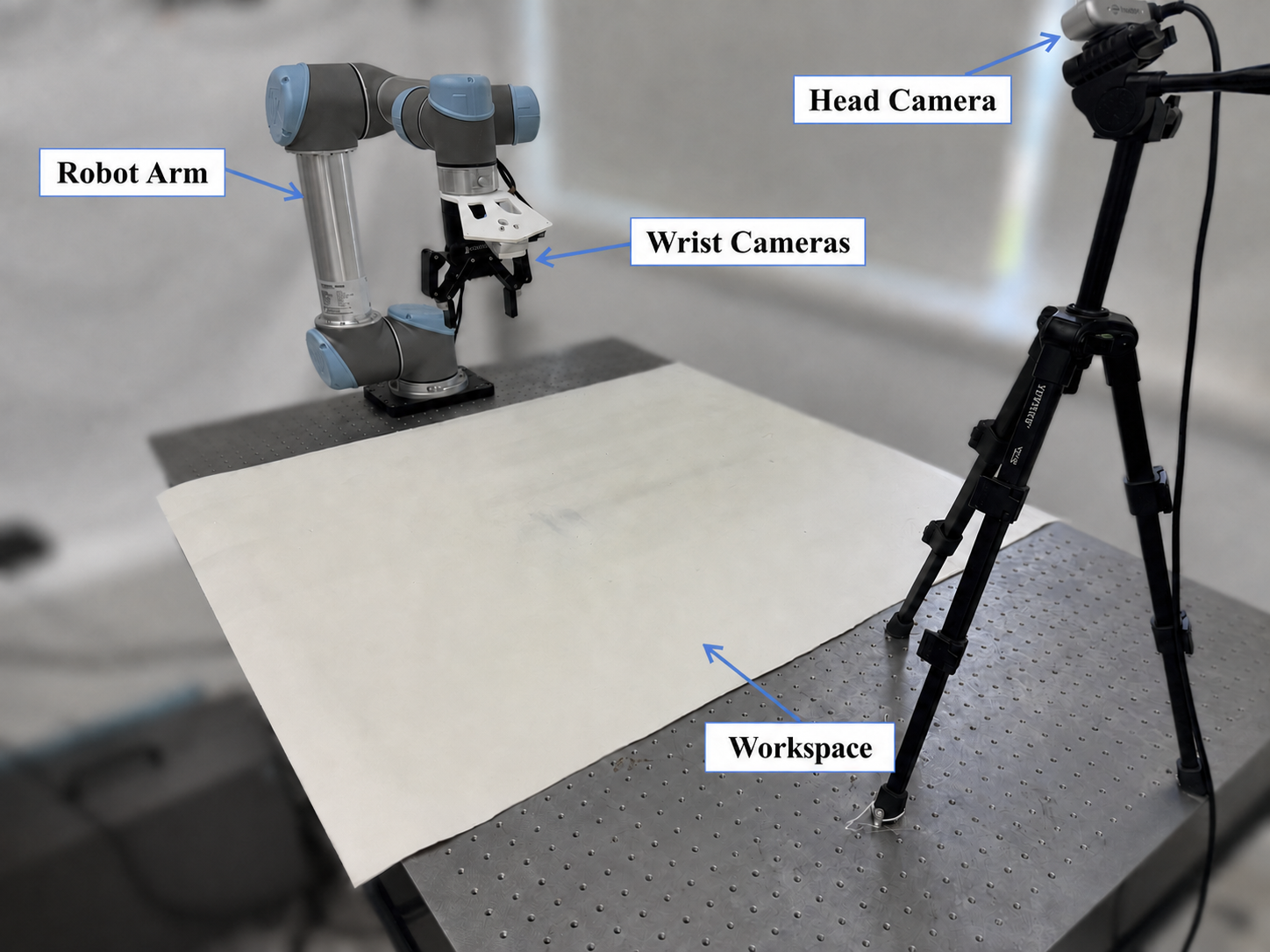}
    \caption{\textbf{Real-robot platform.} The dual-view perception system
    combines global task context with local manipulation observations.}
    \label{supp:fig:supp_hardware}
\end{figure}

\begin{table}[!htbp]
    \centering
    \small
    \caption{\textbf{Robot, camera, and control settings.}
    Hardware, observation, calibration, control, and safety
    configurations used in all real-world experiments.}
    \label{supp:tab:supp_hardware}
    \setlength{\tabcolsep}{4pt}
    \begin{tabularx}{\linewidth}{@{}p{0.34\linewidth}Y@{}}
        \toprule
        Setting & Value \\
        \midrule

        Robot / gripper &
        UR5 / Robotiq 2F-85 \\

        Cameras &
        2 $\times$ Intel RealSense D435i \\

        Camera placement &
        Fixed third-person and wrist-mounted views \\

        Camera capture &
        $640 \times 480$ RGB at 30 Hz \\

        Policy image resolution &
        $224 \times 224$ RGB with aspect-ratio-preserving resize \\

        Hand--eye calibration &
        Target-based calibration; extrinsics fixed across trials \\

        Base--world calibration &
        Target-based rigid-frame alignment; fixed across trials \\

        Camera synchronization &
        Software synchronization using nearest timestamps \\

        Robot control mode / frequency &
        Joint-position control at 30 Hz \\

        Action space &
        6 joint positions and one binary gripper command \\

        Joint safety limits &
        Controller-enforced position, velocity, and acceleration limits \\

        Gripper encoding &
        $0$ if raw command $<50$; $1$ otherwise \\

        Episode horizon &
        1500 control steps \\

        Safety termination &
        Emergency stop or risk of collision, joint-limit violation,
        or workspace exit \\
        \bottomrule
    \end{tabularx}
\end{table}

\subsection{Eight-task benchmark}

All eight tasks are used to evaluate sim--real behavioral consistency. Four
representative tasks are additionally used for policy-refinement experiments.
In-distribution (ID) conditions follow the configurations covered by the
collected demonstrations. Out-of-distribution (OOD) conditions correspond to
deployment-derived failure configurations not covered by those demonstrations.

\begin{table}[!htbp]
    \centering
    \small

    \caption{\textbf{Eight-task benchmark.} Language instructions,
    objects, binary success criteria, episode horizons, and numbers
    of real-world demonstrations used to train the evaluated
    checkpoints.}
    \label{supp:tab:supp_tasks}
    \setlength{\tabcolsep}{5pt}
    \begin{tabularx}{\linewidth}{@{}p{0.16\linewidth}p{0.23\linewidth}p{0.16\linewidth}Yp{0.10\linewidth}@{}}
        \toprule
        Task & Language instruction & Objects & Success criterion &
        Horizon (steps) \\
        \midrule

        Pick Fruits &
        Place the fruit in the basket. &
        fruit, basket &
        Fruit remains stably contained in the basket &
        1500 \\

        Place Cup on Coaster &
        Place the cup on the coaster. &
        cup, coaster &
        Cup remains stably supported by the coaster &
        1500 \\

        Hang Cup &
        Hang the cup on the rack. &
        cup, rack &
        Cup remains suspended from the rack after release &
        1500 \\

        Place Cup in Bowl &
        Place the cup in the bowl. &
        cup, bowl &
        Cup remains stably contained in the bowl &
        1500 \\

        Stack Blocks &
        Stack one block on the other. &
        two blocks &
        Upper block remains stably supported by the lower block &
        1500 \\

        Insert Cylinder into Board &
        Insert the cylinder into the board. &
        cylinder, insertion board &
        Cylinder remains inserted in the designated hole &
        1500 \\

        Place Block in Drawer &
        Put the block in the drawer and close it. &
        block, drawer &
        Block is inside and the drawer is fully closed &
        1500 \\

        Stack Bowls &
        Stack one bowl on the other. &
        two bowls &
        Upper bowl remains stably supported by the lower bowl &
        1500 \\
        \bottomrule
    \end{tabularx}
    
\end{table}

\begin{figure}[!htbp]
    \centering
    \includegraphics[width=0.97\textwidth]{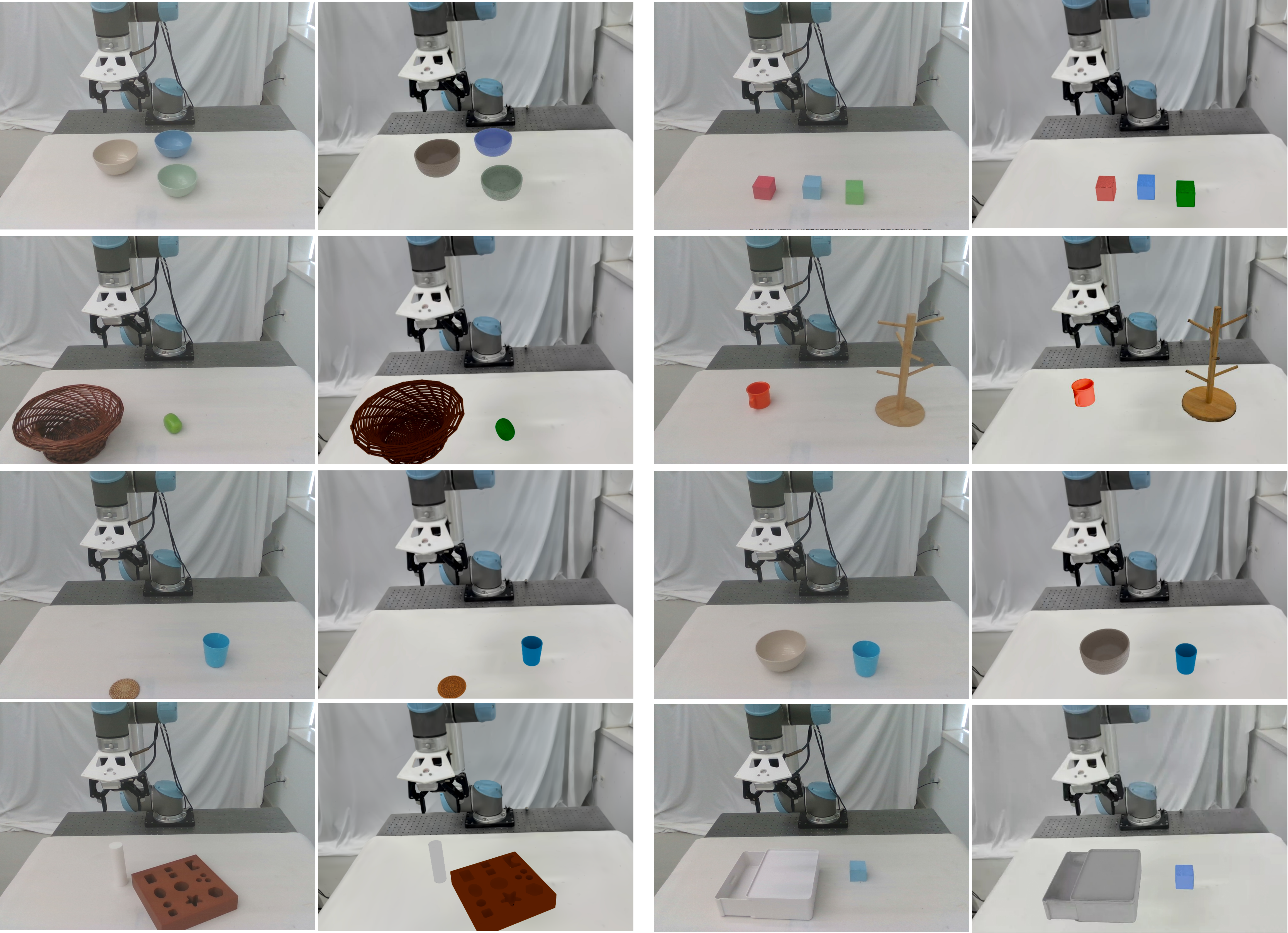}
    \caption{\textbf{Real--simulation paired views.} The reconstructed
    environments preserve the task-relevant objects, spatial relations, and
    viewpoints across all eight tasks.}
    \label{supp:fig:supp_all_tasks}
\end{figure}

\subsection{ID and OOD distributions}

For each task, we evaluate all methods under an in-distribution (ID) setting
and an out-of-distribution (OOD) setting. ID initializations follow the nominal
task distribution used during policy pre-training and standard deployment.
OOD initializations expand the object-pose and distractor-placement ranges
while preserving task semantics and physical feasibility.

All evaluation initializations are pre-generated with fixed random seeds before
running any method. The same initial configurations are reused for all methods
within each task and evaluation setting. In real-world evaluation, we use the
same reset protocol and trial budget for all methods. In simulation, the same
seeded environments are replayed for every method.

Robot initialization, camera extrinsics, illumination, textures, background,
and simulator physical parameters are held fixed across ID and OOD evaluation
unless otherwise specified. For articulated-object tasks, the articulation
state is fixed to the task-specific nominal initial state. Table~\ref{supp:tab:supp_id_ood}
reports the task-specific ID and OOD ranges.

\begin{table}[!htbp]
    \centering
    \small

    \caption{\textbf{ID and OOD evaluation distributions.}
    Positions are expressed in the robot workspace frame.
    OOD cases use deployment-derived failure orientations
    outside the ID yaw range while satisfying the same
    reachability and collision constraints.}
    \label{supp:tab:supp_id_ood}
    \setlength{\tabcolsep}{5pt}
    \begin{tabularx}{\linewidth}{@{}p{0.07\linewidth}p{0.24\linewidth}p{0.23\linewidth}Yp{0.07\linewidth}p{0.07\linewidth}@{}}
\toprule
Setting & Translation range & Yaw range & Feasibility constraint & Trials per task & Shared init. \\
\midrule
ID &
$x\sim\mathcal{U}(0.02,0.88)$ m,\newline
$y\sim\mathcal{U}(-0.185,0.957)$ m,\newline
$z=z_0$
& $\psi\sim\mathcal{U}(-\pi/6,\pi/6)$
& Reachable and collision-free & 20 & Yes \\
\addlinespace
OOD & Same feasible workspace &
$\psi\in[-\pi/3,-\pi/6)$\newline
$\phantom{\psi\in{}}\cup(\pi/6,\pi/3]$
& Reachable and collision-free & 20 & Yes \\
\bottomrule
\end{tabularx}
\medskip
\par\noindent\textit{Task-specific randomized entities and deployment-derived failure variables}\par\smallskip
\begin{tabularx}{\linewidth}{@{}p{0.20\linewidth}YYp{0.13\linewidth}@{}}
\toprule
        Task & Randomized entities &
        Deployment-derived OOD variable &
        Source \\
        \midrule
        Pick Fruits &
        Fruit and basket poses &
        Fruit--basket relative pose &
        Deployment \\

        Place Cup on Coaster &
        Cup and coaster poses &
        Cup--coaster relative pose &
        Deployment \\

        Hang Cup &
        Cup pose &
        Cup--rack relative pose &
        Deployment \\

        Place Cup in Bowl &
        Cup and bowl poses &
        Cup--bowl relative pose &
        Deployment \\

        Stack Blocks &
        Initial poses of both blocks &
        Inter-block displacement and yaw &
        Deployment \\

        Insert Cylinder into Board &
        Cylinder and board poses &
        Cylinder--hole relative pose &
        Deployment \\

        Place Block in Drawer &
        Block pose and drawer state &
        Block--drawer pose and opening &
        Deployment \\

        Stack Bowls &
        Initial poses of both bowls &
        Inter-bowl displacement and yaw &
        Deployment \\
        \bottomrule
    \end{tabularx}
    
\end{table}
\subsection{Success criteria and evaluation protocol}

Real-world performance is measured over 20 trials per task under both ID and
OOD conditions. Simulation performance is measured using one rollout in each
of 100 parallel environments. All methods share the same initial configurations,
trial budget, task horizon, and binary success criteria.

A trial is counted as successful only when the task-specific goal condition is
satisfied at the end of the episode and remains stable after robot release. For
pick-and-place tasks, the manipulated object must be placed inside the target
region without toppling or leaving the workspace. For stacking tasks, all bowls
must remain stacked after release. For drawer tasks, the target object must be
placed inside the drawer and the drawer state must remain within the
task-defined valid range.

Real-world outcomes are judged by a blinded human evaluator from recorded
trial videos according to the predefined binary success criteria. The evaluator
is not shown the method identity during scoring. Ambiguous outcomes are counted
as failures unless the success condition is clearly satisfied. For each task and evaluation setting, we report the success rate over 20 trials.

\subsection{Baselines and budget matching}

\paragraph{Targeted BC.}
Targeted BC relies on human diagnosis of the observed failures and collects
corrective demonstrations in the corresponding real-world failure
distribution. It is allocated one hour of end-to-end real-world
data-preparation time, including scene initialization, object rearrangement,
environment resets, and preparation for repeated collection attempts. The
allocated time excludes subsequent policy training. Targeted BC uses the same
base policy, LoRA configuration, and SFT procedure as \FfourR.

\paragraph{RLinf-Co.}
RLinf-Co~\citep{shi2026beyond} performs sim--real co-training followed by
simulation-based RL. To isolate the effect of failure-aware
randomization, RLinf-Co and \FfourR use the same policy
architecture, real-world training data, simulator assets,
optimization hyperparameters, update schedule, number of
parallel environments, and rollout budget. They differ only
in the construction of the simulated training distribution:
RLinf-Co applies broad, task-level domain randomization,
whereas \FfourR concentrates randomization around failure
modes diagnosed from real-world deployments.

\paragraph{Budget matching.}
We match all methods by an end-to-end data-preparation
wall-clock budget rather than by trajectory count. Targeted
BC is allocated one hour for real-world data collection,
including scene initialization, object rearrangement,
environment resets, and preparation for repeated collection
attempts. RLinf-Co and \FfourR are each allocated the same
one-hour budget: 50 minutes for method-specific preparation
and 10 minutes for parallel simulated rollout collection.
For \FfourR, preparation includes failure diagnosis,
scene and object reconstruction, simulator integration, and
validation. For RLinf-Co, the same allowance covers simulator
and object-asset construction, configuration of broad domain
randomization, integration, and validation. Scene
reconstruction typically requires 25--30 minutes, while
preparing a new object asset requires 5--10 minutes; the
remaining time is used for integration and validation.

The reconstructed scene is a reusable asset and only needs
to be updated when the physical workspace changes. When only
the manipulated object changes, the scene is reused and only
a new object asset is prepared. Because task horizons and
parallel collection produce different numbers of
trajectories, we control for data-preparation time rather
than trajectory count.

\begin{table}[!htbp]
    \centering
    \small
    \caption{\textbf{Budget and cost breakdown.} Initial-cycle and amortized
    costs under the matched comparison.}
    \label{supp:tab:supp_budget}
    \setlength{\tabcolsep}{4pt}
    \begin{tabularx}{\linewidth}{@{}p{0.23\linewidth}Y Y Y Y Y Y@{}}
        \toprule
        Method & Human corrective data & Scene setup & Object setup &
        Sim collection & Training & Reusable assets \\
        \midrule

        Targeted BC & 60 min & 0 & 0 & 0 & 15h & --- \\
        RLinf-Co & 0 & shared 25--30 min & 5--10 min & 10 min & 10-14h & Yes \\
        \FfourR & 0 & shared 25--30 min & 5--10 min & 10 min & 10-14h & Yes \\
        \midrule
        Later \FfourR cycle, unchanged workspace/object &
        0 & 0 & 0 & 10 min & 10-14h & Reused \\
        \bottomrule
    \end{tabularx}
    
\end{table}

\begin{table}[!htbp]
    \centering
    \small

    \caption{\textbf{Per-task Targeted BC data collection.}
    Failed attempts are excluded from the training data.
    Trajectory length is measured as the number of transitions
    recorded at 30\,Hz.}
    \label{supp:tab:supp_targeted_bc}
    \begin{tabularx}{\linewidth}{@{}Ycccc@{}}
        \toprule
        Task & Collection time & Successful demos & Failed attempts
        & Mean length \\
        \midrule

        Pick Fruits
        & 1.0\,h & 48 & 2 & 570 \\

        Place Cup on Coaster
        & 1.0\,h & 42 & 3 & 630 \\

        Stack Bowls
        & 1.0\,h & 24 & 3 & 1,740 \\

        Place Block in Drawer
        & 1.0\,h & 35 & 2 & 860 \\

        \bottomrule
    \end{tabularx}
\end{table}

\FloatBarrier
\section{Additional Quantitative Results}
\label{supp:sec:additional-results}

\subsection{Complete real-world results}

Table~\ref{supp:tab:supp_main_counts} reports the count form of the main real-world
results. Each entry is the number of successful trials out of 20. Macro
averages are computed by averaging the four task success rates with equal task
weight. 

\begin{table}[!htbp]
    \centering
    \small
    \caption{\textbf{Complete real-world results.} Successful trials out of 20
    under ID and deployment-derived OOD conditions.}
    \label{supp:tab:supp_main_counts}
    \setlength{\tabcolsep}{3.5pt}
    \begin{adjustbox}{max width=\textwidth}
    \begin{tabular}{lcccccccccc}
        \toprule
        \multirow{2}{*}{Method} &
        \multicolumn{4}{c}{ID} & \multirow{2}{*}{ID Avg.} &
        \multicolumn{4}{c}{OOD} & \multirow{2}{*}{OOD Avg.} \\
        \cmidrule(lr){2-5}\cmidrule(lr){7-10}
        & Pick & Cup & Bowls & Drawer & & Pick & Cup & Bowls & Drawer & \\
        \midrule
        Base & 14/20 & 12/20 & 10/20 & 14/20 & 62.50\% &
        8/20 & 7/20 & 6/20 & 0/20 & 26.25\% \\
        Targeted BC & 20/20 & 19/20 & 16/20 & 18/20 & 91.25\% &
        17/20 & 16/20 & 12/20 & 12/20 & 71.25\% \\
        RLinf-Co & 19/20 & 17/20 & 16/20 & 18/20 & 87.50\% &
        18/20 & 14/20 & 14/20 & 16/20 & 77.50\% \\
        \textbf{\FfourR} & \textbf{20/20} & \textbf{19/20} &
        \textbf{17/20} & \textbf{19/20} & \textbf{93.75\%} &
        \textbf{20/20} & \textbf{19/20} & \textbf{16/20} &
        \textbf{17/20} & \textbf{90.00\%} \\
        \bottomrule
    \end{tabular}
    \end{adjustbox}
\end{table}

\FfourR improves OOD success by 18.75 percentage points over Targeted BC and
12.5 points over RLinf-Co. The largest distinction is not simply additional
optimization: RLinf-Co uses an identical optimization and rollout budget but
spreads interaction across broad variations. Failure diagnosis instead
concentrates \FfourR's data generation and RL interaction on
deployment-relevant weaknesses.

\subsection{Full sim--real consistency results}

For each of eight tasks, we train policies using 10, 30, and 50 real-world
demonstrations, yielding 24 checkpoints. Each checkpoint is evaluated under
matched configurations and success criteria in simulation and the real world.
The resulting success rates exhibit a strong positive correlation
($r=0.7625$, $p=1.49\times10^{-5}$), showing that the reconstructed environments
preserve both broad task difficulty and performance changes induced by
additional data. Table~\ref{supp:tab:supp_sim_real_24} reports all checkpoint
values.

\begin{table}[!htbp]
    \centering
    \small
    \caption{\textbf{Complete sim--real consistency results.}
    Raw success rates for all 24 checkpoints. The sim--real gap
    $|\Delta|$ is reported in percentage points (pp).}
    \label{supp:tab:supp_sim_real_24}
    \setlength{\tabcolsep}{5pt}
    \begin{tabularx}{\linewidth}{@{}p{0.25\linewidth}YYYYY@{}}
        \toprule
        Task & Demonstra-\newline tions & Real success (\%) & Sim success (\%) &
        $|\Delta|$ (pp) & Trials (real/sim) \\
        \midrule

        Pick Fruits & 10 & 20 & 25 & 5 & 20/100 \\
        Pick Fruits & 30 & 50 & 47 & 3 & 20/100 \\
        Pick Fruits & 50 & 100 & 77 & 23 & 20/100 \\
        \midrule
        Put Cup on Coaster & 10 & 20 & 17 & 3 & 20/100 \\
        Put Cup on Coaster & 30 & 70 & 60 & 10 & 20/100 \\
        Put Cup on Coaster & 50 & 95 & 76 & 19 & 20/100 \\
        \midrule
        Hang Cup & 10 & 10 & 0 & 10 & 20/100 \\
        Hang Cup & 30 & 60 & 10 & 50 & 20/100 \\
        Hang Cup & 50 & 80 & 20 & 60 & 20/100 \\
        \midrule
        Place Cup in Bowl & 10 & 80 & 25 & 55 & 20/100 \\
        Place Cup in Bowl & 30 & 100 & 55 & 45 & 20/100 \\
        Place Cup in Bowl & 50 & 100 & 62 & 38 & 20/100 \\
        \midrule
        Stack Blocks & 10 & 10 & 4 & 6 & 20/100 \\
        Stack Blocks & 30 & 50 & 19 & 31 & 20/100 \\
        Stack Blocks & 50 & 60 & 26 & 34 & 20/100 \\
        \midrule
        Insert Cylinder into Board & 10 & 0 & 0 & 0 & 20/100 \\
        Insert Cylinder into Board & 50 & 5 & 7 & 2 & 20/100 \\
        Insert Cylinder into Board & 100 & 20 & 10 & 10 & 20/100 \\
        \midrule
        Place Block in Drawer & 10 & 40 & 36 & 4 & 20/100 \\
        Place Block in Drawer & 30 & 70 & 65 & 5 & 20/100 \\
        Place Block in Drawer & 50 & 100 & 95 & 5 & 20/100 \\
        \midrule
        Stack Bowls & 10 & 30 & 37 & 7 & 20/100 \\
        Stack Bowls & 30 & 50 & 78 & 28 & 20/100 \\
        Stack Bowls & 50 & 80 & 51 & 29 & 20/100 \\
        \bottomrule
    \end{tabularx}
\end{table}

\subsection{Multi-round closed-loop refinement}

We conduct three consecutive refinement cycles on Stack Bowls and Pick Fruits.
After each cycle, the updated policy is redeployed and newly observed failures
are fed into the next diagnosis and refinement cycle. Stack Bowls improves from
$30\%$ to $80\%$, $90\%$, and $95\%$. Pick Fruits improves from $40\%$ to
$100\%$ in the first cycle and maintains this performance in later evaluations.
These results show that \FfourR can address residual failures while retaining
capabilities acquired in earlier rounds. Table~\ref{supp:tab:supp_closed_loop}
details each cycle.

\begin{table}[!htbp]
    \centering

    \caption{\textbf{Per-cycle closed-loop refinement.} Diagnosed failures,
    randomized variables, new data, and real-world success.}
    \label{supp:tab:supp_closed_loop}
    \setlength{\tabcolsep}{3.5pt}
    \begin{adjustbox}{max width=\textwidth}
    \begin{tabular}{ccccc}
        \toprule
        Task & Cycle & Success & Newly observed failure & Assets reused \\
        \midrule
        Stack Bowls & 0 & 30\% & Initial deployment failures & \NA \\
        Stack Bowls & 1 & 80\% & Object Order & Yes \\
        Stack Bowls & 2 & 90\% & Object Placement & Yes \\
        Stack Bowls & 3 & 95\% & Illumination & Yes \\
        \midrule
        Pick Fruits & 0 & 40\% & Initial deployment failures & \NA \\
        Pick Fruits & 1 & 100\% & Object Placement & Yes \\
        Pick Fruits & 2 & 100\% & \NA & Yes \\
        Pick Fruits & 3 & 100\% & \NA & Yes \\
        \bottomrule
    \end{tabular}
    \end{adjustbox}
\end{table}

\subsection{Cross-task capability retention}

To evaluate whether F4R supports continual improvement beyond one-shot refinement, we conduct two consecutive deployment--refinement cycles. After each cycle, the refined policy is redeployed, and newly observed failures are used to guide the next round of reconstruction and training. 

\begin{figure}[!htbp]
    \centering
    \includegraphics[width=0.72\linewidth]{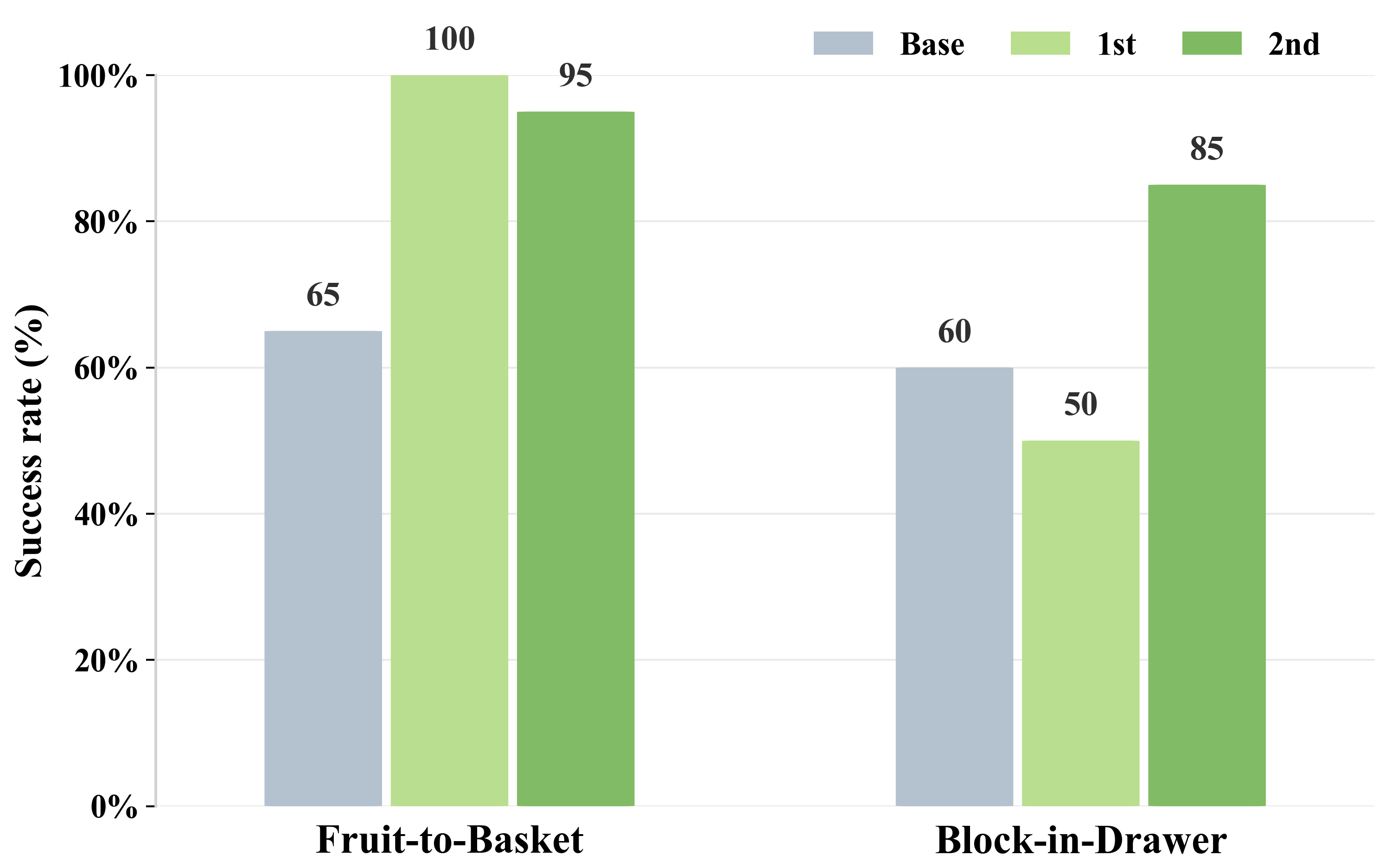}
    \caption{Evaluation on multi tasks.}
    \label{supp:fig:closed_loop}
\end{figure}

As shown in Figure~\ref{supp:fig:closed_loop}, the success rate on \textit{Fruit-to-Basket} increases from 65\% to 100\% after the first cycle and remains near saturation at 95\% after the second. On \textit{Cup-on-Coaster}, performance initially decreases from 60\% to 50\%, indicating that a single targeted update may expose or introduce failure modes not covered by the current training distribution. Feeding these failures into the second cycle raises the success rate to 90\%. Overall, the average success rate increases from 62.5\% to 92.5\% after two cycles, demonstrating that F4R can repeatedly convert deployment feedback into further policy improvement.

\FloatBarrier
\section{Qualitative Results and Limitations}
\label{supp:sec:qualitative}

\subsection{Failure cases}

Table~\ref{supp:tab:supp_limitations} summarizes these limitations and possible
mitigations.

\begin{table}[!htbp]
    \centering
    \small
    \caption{\textbf{Current limitations and possible mitigations.}}
    \label{supp:tab:supp_limitations}
    \begin{tabularx}{\linewidth}{@{}p{0.14\linewidth}YYY@{}}
        \toprule
        Component & Failure mode & Consequence & Possible mitigation \\
        \midrule

        Diagnosis & Subtle viewpoint or illumination change &
        Incorrect causal attribution & Longer temporal memory and explicit view normalization \\
        Diagnosis & Multiple simultaneous failures &
        Ambiguous earliest cause & Causal multi-hypothesis diagnosis \\
        Reconstruction & Transparent, reflective, or thin objects &
        Incomplete geometry/depth & Specialized sensing or material-aware reconstruction \\
        Reconstruction & Ambiguous articulation &
        Incorrect joint axis/limits & Active articulation probing \\
        Simulation & Friction/contact mismatch &
        Sim--real execution gap & Online parameter identification \\
        Simulation & Deformable objects or dynamic scenes &
        Invalid rigid-scene assumption & Deformable/dynamic reconstruction \\
        Refinement & Sparse-reward exploration failure &
        Insufficient PPO improvement & Better initialization or generic progress rewards \\
        Refinement & Overly local failure distribution &
        Limited broader robustness & Adaptive mixture with broad randomization \\
        Continual loop & Growing failure memory &
        Increasing sampling/training cost & Failure clustering and curriculum management \\
        \bottomrule
    \end{tabularx}
    
\end{table}

\subsection{Scope and safety}

Our current evaluation focuses on tabletop manipulation with predominantly
rigid objects. The method has not yet been validated on deformable objects,
dynamic scenes, mobile manipulation, or long-horizon assembly. Scaling to many
objects and accumulated failures also requires strategies for merging,
prioritizing, and retiring failure distributions.

During real-robot deployment, a human operator remains available for emergency stop and hardware safety, but does not diagnose failures or provide corrective
demonstrations for \FfourR. Rollouts terminate on timeout, communication loss,
detected collision, or other abnormal conditions. By shifting repeated
failure correction to simulation, \FfourR reduces the number of potentially
risky real-world retries.

\subsection{Reproducibility}

\paragraph{Software and model versions.}
We use $\pi_{0.5}$ as the base VLA policy and GPT-5.5 as the failure-diagnosis backend. The simulation environment is built with Isaac Sim 5.1 and Isaac Lab 2.3.2. Training uses CUDA 12.4, PyTorch 2.6.0, Transformers 4.40.1, and Ray 2.54.0, while the real-robot system runs ROS Noetic. All experiments use a fixed random seed of 42. Detailed training hyperparameters are provided in Table~\ref{supp:tab:supp_training_hyperparameters}.

\end{document}